\documentclass[preprint,12pt]{elsarticle}

\usepackage{soul, color}
\usepackage{graphicx}
\usepackage{adjustbox}
\usepackage{booktabs,makecell}
\usepackage{makecell}
\usepackage{tabularx}
\usepackage{threeparttable, tablefootnote}
\usepackage{amsmath}

\usepackage{newunicodechar}

\usepackage[colorlinks,allcolors=black]{hyperref}
\usepackage{hyperref}
\hypersetup{colorlinks,allcolors=black}

\usepackage{graphicx}
\usepackage{multirow}
\usepackage[dvipsnames]{xcolor}
\usepackage{soul}
\usepackage{color}
\usepackage{url}
\usepackage{float}
\usepackage{boldline}
\usepackage{xcolor,colortbl}
\usepackage{tikz}

\usepackage{pifont}

\usepackage{multirow}
\usepackage{graphicx}
\usepackage{amssymb}
\usepackage{soul}
\usepackage[normalem]{ulem}
\usepackage{xcolor}

\usepackage{xspace}
\usepackage{tabularx}

\journal{}

\begin{document}

\begin{frontmatter}


\title{Hybrid Ensemble Approaches: Optimal Deep Feature Fusion and Hyperparameter-Tuned Classifier Ensembling for Enhanced Brain Tumor Classification}

\author{Zahid Ullah$^{1}$, Dragan Pamucar$^{2,3,4}$, Jihie Kim$^{1,*}$}

\address{%
$^{1}$ \quad Department of Computer Science and Artificial Intelligence, Dongguk University, Seoul 04620, Republic of Korea \\  
$^{2}$ \quad Department of Operations Research and Statistics, Faculty of Organizational Sciences, University of Belgrade, Belgrade, Serbia \\ 
$^{3}$ \quad Department of Industrial Engineering $\&$ Management, Yuan Ze University, Taoyuan City 320315, Taiwan \\
$^{4}$ \quad Department of Applied Mathematical Science, College of Science and Technology, Korea University, Sejong 30019, Republic of Korea  }

\begin{abstract}

Magnetic Resonance Imaging (MRI) is widely recognized as the most reliable tool for detecting tumors due to its capability to produce detailed images that reveal their presence. However, the accuracy of diagnosis can be compromised when human specialists evaluate these images. Factors such as fatigue, limited expertise, and insufficient image detail can lead to errors. For example, small tumors might go unnoticed, or overlap with healthy brain regions could result in misidentification. To address these challenges and enhance diagnostic precision, this study proposes a novel double ensembling framework, consisting of ensembled pre-trained deep learning (DL) models for feature extraction and ensembled fine-tuned hyperparameter machine learning (ML) models to efficiently classify brain tumors. Specifically, our method includes extensive preprocessing and augmentation, transfer learning concepts by utilizing various pre-trained deep convolutional neural networks and vision transformer networks to extract deep features from brain MRI, and fine-tune hyperparameters of ML classifiers. Our experiments utilized three different publicly available Kaggle MRI brain tumor datasets to evaluate the pre-trained DL feature extractor models, ML classifiers, and the effectiveness of an ensemble of deep features along with an ensemble of ML classifiers for brain tumor classification. Our results indicate that the proposed feature fusion and classifier fusion improve upon the state of the art, with hyperparameter fine-tuning providing a significant enhancement over the ensemble method. Additionally, we present an ablation study to illustrate how each component contributes to accurate brain tumor classification.


\end{abstract}

\end{frontmatter}


\section{Introduction}
\label{intro}
A brain tumor is an abnormal, unchecked proliferation of cells within the brain and is among the most dangerous diseases of the nervous system. The National Brain Tumor Society reports that each year, roughly 400,000 people worldwide are diagnosed with a brain tumor, and about 120,000 of these individuals die annually—a number that continues to rise. In adults, the most prevalent type of primary brain tumor is glioma, which can inflict severe damage on the central nervous system. There are four Grades of brain tumor, such as 1 and 2 indicating less aggressive tumors (such as meningiomas), whereas grades 3 and 4 include the more aggressive forms (like gliomas). Clinically, meningiomas, pituitary tumors, and gliomas represent roughly 15\%, 15\%, and 45\% of brain tumor cases, respectively.

In all of the available imaging modalities, MRI is often preferred for brain tumor assessment due to its non-invasive nature and absence of ionizing radiation, providing detailed two-dimensional and three-dimensional information about tumor size, shape, type, and location \cite{ullah2022cascade}. However, manually analyzing these images can be both time-intensive and error-prone, especially with a high volume of patients \cite{popuri20123d}. To mitigate this issue, developing automated computer-aided diagnosis systems is essential to reduce the workload associated with classifying and diagnosing brain MRI scans, thereby assisting radiologists and physicians. Unfortunately, the automatic brain tumor classification is complex due to issues like low-quality images, insufficient training data, inadequate image attributes, and inaccurate tumor location can impede the classification process \cite{ullah2020hybrid}.

Extensive research has been dedicated to creating an automated, highly reliable system for classifying brain tumors. Nonetheless, the task remains difficult due to significant differences in shape, texture, and contrast both across different cases and within the same case. Conventional machine learning (ML) methods depend on features that are manually crafted, which limits their flexibility and overall robustness. In contrast, deep learning (DL) approaches automatically learn relevant features from the data, resulting in substantially improved performance. DL methods typically demand extensive labeled datasets, which can be difficult to obtain. To address this challenge, our study introduces a hybrid approach that leverages twelve pre-trained deep convolutional neural networks (CNNs) and thirteen pre-trained vision transformer-based (ViTs) models as feature extractors to capture robust and distinctive attributes from brain MRI scans. These features are subsequently input into ML classifiers to differentiate between tumorous MRI and normal MRI. 

To explore the advantages of integrating attributes from various pre-trained DL  models, we developed a novel feature ensemble approach for classifying brain tumor using different versions such as simple version, with normalization and PCA, with SMOTE, and using the combination of normalization, PCA, and SMOTE as discussed in section \ref{abstu}. Our approach introduces an innovative feature evaluation and selection mechanism, where deep features extracted from twenty-five pre-trained DL models are assessed using 9 distinct ML classifiers and selected based on a proposed selection criterion. In our framework, the top-2 or top-3 deep feature extractor models are combined to form a synthetic feature. This concatenation process merges diverse information from multiple pretrained DL architectures, resulting in a more robust and discriminative feature representation compared to using features from a single pretrained DL model. The ensemble of deep features is then input into various ML classifiers to generate the final prediction. Unlike most previous studies, which primarily relied on traditional feature extraction methods, our approach integrates diverse DL features for improved classification performance.

Specifically, we propose a novel double ensemble technique where, on the one hand, we combine the top-2 as well as the top-3 deep feature extractor models and feed those DL-based extracted features to 9 different ML classifiers to classify the MRI into normal or abnormal. On the other hand, we also ensembled the top-2 or top-3 ML classifiers and presented individually the top-5 pre-trained DL model features to predict the output. Hence, this ensembling technique significantly enhances the performance, as illustrated in our experimental results.
The contributions of our research study are as follows:

\begin{itemize}
    \item We adopted transfer learning techniques to acquire more discriminative characteristics with adequate information and feed them to ML classifiers to accurately classify the brain tumor. 
    \item We conducted hyperparameter tuning using grid search to optimize ML model performance, significantly improving accuracy.
 \item We highlighted the significance of feature extraction and selection methods in improving the performance and efficiency of ML models for medical imaging applications.
    \item Our proposed model consists of four steps: first, deep features are extracted using pre-trained DL models to ensure meaningful information extraction and improved generalization. Second, the top three most effective features are identified by fine-tuning multiple ML models for our specific task. Third, these features are integrated to construct an ensemble model aimed at achieving state-of-the-art (SOTA) performance in brain tumor classification using brain MRI images. Finally, the top three fine-tuned ML models are ensembled to improve the accuracy of brain tumor classification.

\end{itemize}

The code implementation of our approach will be openly shared after publication to promote reproducibility and facilitate further research. Code Link: \url{https://github.com/Zahid672/Brain-Tumor-Classification}.

The remaining paper is organized as follows: Section \ref{related} presents a comprehensive review of related work to contextualize our research. Next, in Section \ref{pm} we detail our proposed methodology, including the dataset and the main architecture. We then describe our experimental setup and implementation details in section \ref{experimental}. Section \ref{results} consists of the results, whereas section \ref{discussion} presents the discussion. Finally, in section \ref{con}, we present the conclusion.

\section{Related Work}\label{related}
\subsection{Feature Extraction and Classification}
In this study, DL-based feature extraction and traditional ML classifiers were employed for brain tumor classification using MRI. Two types of models were utilized for feature extraction: CNNs and ViTs.

\subsection{Feature Extraction with CNNs}
CNN-based models are widely recognized for their ability to extract hierarchical spatial features from images. These models leverage convolutional layers to capture local patterns, which are essential for identifying relevant features in medical images like MRI scans. Pretrained CNN architectures, such as ResNet, DenseNet, VGG, AlexNet, ResNext, ShuffleNet, MobileNet, and MnasNet, have been employed to capture deep features from the MRI data. 


\subsection{Feature Extraction with ViTs}
ViT-based models represent a more recent approach to feature extraction, leveraging self-attention methods to identify global relationships within an image. Unlike CNNs, which rely on local receptive fields, ViTs process images as sequences of patches and learn contextual dependencies across the entire image. Pretrained ViT architectures, such as ViT-Base Transformer, were employed to extract feature vectors from the MRI images, providing complementary information to the ViT-based features.

\subsection{Classification with Machine Learning Models}
Once the feature vectors were extracted, they were fed into traditional ML classifiers for brain tumor classification, such classifiers were used to predict the presence of a brain tumor. These classifiers were trained and evaluated on the extracted feature sets to achieve robust and accurate predictions.

Over the past decade, DL techniques have significantly advanced the classification of brain MRI images \cite{saleh2020brain,waghmare2021brain,aamir2022deep}. Unlike conventional approaches that depend on manually designed features, DL models autonomously integrate feature extraction and classification, streamlining the analysis process. While these models often require preprocessing steps, they excel at identifying critical features without explicit human intervention. CNNs, a prominent DL architecture for image analysis, effectively address this issue by serving as automated feature extractors.

In CNNs, the starting layers concentrate on gaining low-level features. Subsequent, deeper layers synthesize these basic features into high-level, domain-specific representations. This hierarchical approach enables CNNs to construct comprehensive feature representations, combining detailed structural information from earlier layers with broader contextual insights from deeper layers, thereby enhancing the accuracy and efficiency of MRI classification tasks.

In recent years, different studies leveraged CNNs for the classification of brain MRI scans, particularly focusing on brain tumor detection and categorization \cite{sandhiya2024deep,agrawal2024comparative,priya2024brain}. For example, the authors in \cite{deepak2019brain} extract features using a pre-trained GoogLeNet model from brain MRI, attaining 98\% classification accuracy across three tumor types. 
 Similarly, Ahmet and Muhammad \cite{ccinar2020detection} modified a pre-trained ResNet-50 model by replacing the last five layers and adding eight new layers, resulting in a classification accuracy of 97.2\%, the highest among the models they tested. 
 Khwaldeh et al. \cite{khawaldeh2017noninvasive} adapted the AlexNet architecture to differentiate normal and cancerous MRI, obtaining 91\% accuracy. Despite these advancements, there is still a need for more robust and practical methods to further improve the accuracy and reliability of brain MRI classifications. Additionally, Saxena et al. \cite{saxena2020predictive} explored transfer learning techniques using different pre-trained models to classify brain tumors. Out of all, the ResNet-50 attained the most accurate results. In \cite{yang2018feature,wicht2017deep} introduced CNN architectures designed to categorize brain tumors. These CNNs utilize convolution and pooling techniques for feature extraction. Their aim was to determine the most effective DL approach for accurately classifying brain MR images. In study \cite{diaz2021deep}, developed a multi-pathway CNN architecture for the automated segmentation of brain tumors, achieving an impressive 97.3\% accuracy, though the training process was notably resource-intensive. Meanwhile, Raja et al. \cite{raja2020brain} suggested a hybrid deep autoencoder method to classify brain tumors, incorporating a Bayesian fuzzy clustering (BFC) technique. First, they use a mean filter to eliminate the noise, followed by the BFC method for tumor segmentation. They then extracted key features. Ultimately, their hybrid DAE approach delivered high accuracy in classification, though it demanded significant computational time. Sahin et al. \cite{csahin2024multi} present an approach to enhance the ViT model for classifying brain tumors. The study fine-tunes the ViT's architectural parameters, where their research findings suggest that integrating BMO with ViT models can lead to more efficient and accurate brain tumor classification. However, the lack of external validation on diverse datasets raises concerns about the model's generalizability. Wang et al. \cite{wang2024ranmerformer}  proposed a model called RanMerFormer, utilizes a pre-trained ViT as its backbone and incorporates a token merging mechanism to eliminate redundant tokens, thereby significantly improving computational efficiency. Additionally, it employs a randomized vector functional-link as the classification head, which facilitates rapid training and obtains SOTA performance. 
 

 \subsection{Summary}

 In summary, most of the existing studies show that DL techniques deliver significantly greater precision in brain MRI classification than standard ML methods. However, these DL systems require a large quantity of data for training to surpass the performance of traditional ML approaches. Recent publications clearly demonstrate that DL techniques have emerged as a dominant approach in medical image analysis. However, these techniques come with specific limitations that need to be taken into account when addressing brain tumor classification and segmentation. A significant weakness of earlier systems is their focus solely on binary classification (normal versus abnormal) of MR images, while overlooking datasets involving multiple classes \cite{bhuvaji2020brain}. In this work, We developed a fully automated hybrid approach for brain tumor classification that combines two key components: pre-trained DL models to capture deep features from brain MRI, and ML classifiers to accurately determine the type of brain tumor. Furthermore, our work mainly consists of 4 distinct steps, which are unique from previous studies to the best of our knowledge. First, we extracted deep features from MR images using pre-trained DL models. Second, these extracted features are rigorously evaluated using 9 different ML classifiers. Third, we selected the top 2 as well as the top 3 deep feature extractor models identified based on their accuracy across multiple classifiers, and combined them into a deep feature ensemble. This ensemble is then fed into different ML classifiers to produce the final prediction. Lastly, we also explored creating an ensemble of the top two or three ML classifiers by incorporating features from the top five individual pre-trained DL models.
 
\section{Proposed Methodology} 
\label{pm}

\subsection{Overview}
\begin{figure*}[!ht]
     \centering
     \includegraphics[width=1\textwidth]{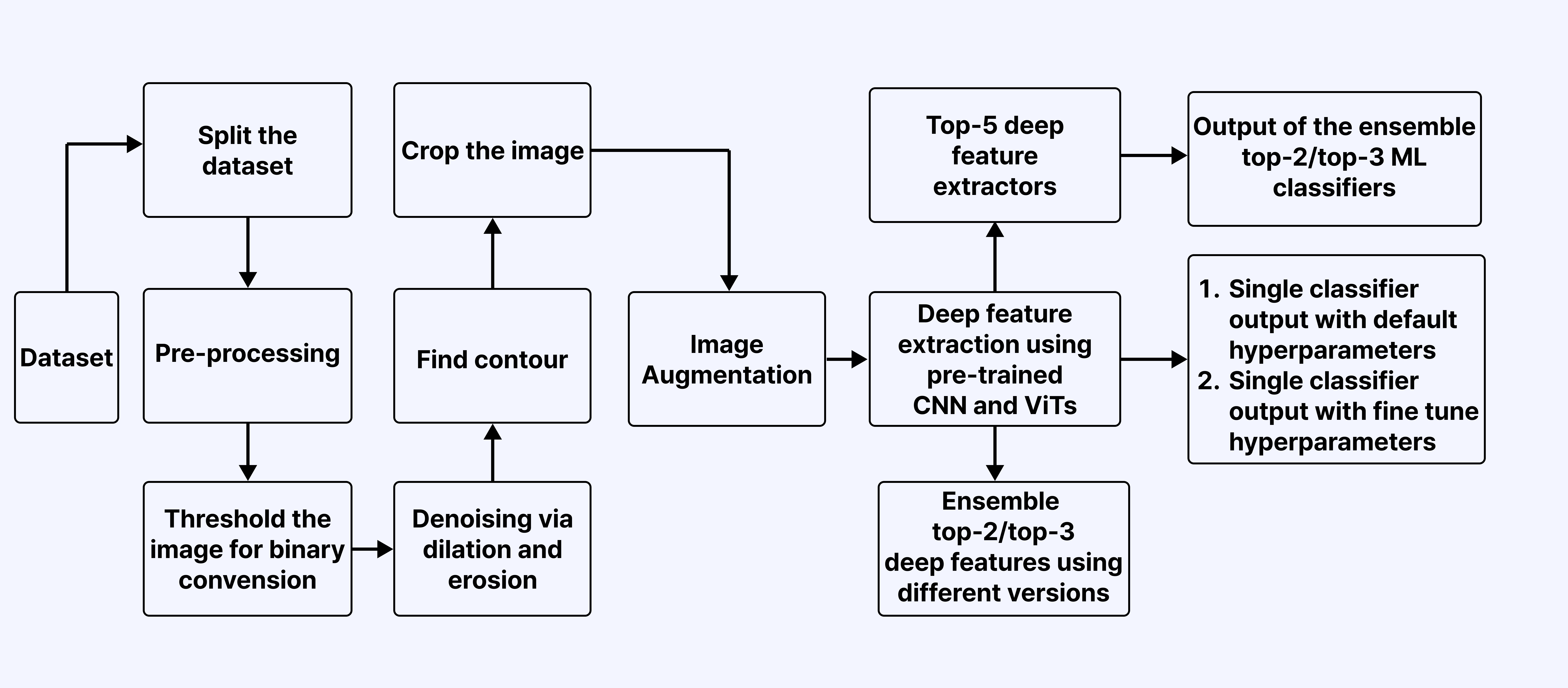}
     \caption{Workflow of our proposed double ensembling framework for brain tumor classification. }
     \label{workflow}
 \end{figure*}

The workflow of this study is depicted in Fig. \ref{workflow}. This section outlines a comprehensive architecture for brain tumor classification using DL techniques. The process begins with preprocessing MR images, including cropping, resizing, and augmentation as shown in section \ref{pp}. We presented these preprocessed images to pre-trained CNN and ViT models for feature extraction as depicted in section \ref{feaExt}. The extracted features undergo evaluation using various ML classifiers as illustrated in section \ref{MLcl}. Based on performance, the top 2 or top 3 deep features are selected and combined in an ensemble module. Similarly, these concatenated features serve as input for ML classifiers to make the final prediction. This approach leverages transfer learning and ensemble techniques to improve the accuracy and robustness of brain tumor classification, as shown in section \ref{abstu}.

\subsection{Datasets}
\begin{figure*}[!ht]
     \centering
     \includegraphics[width=0.9\textwidth]{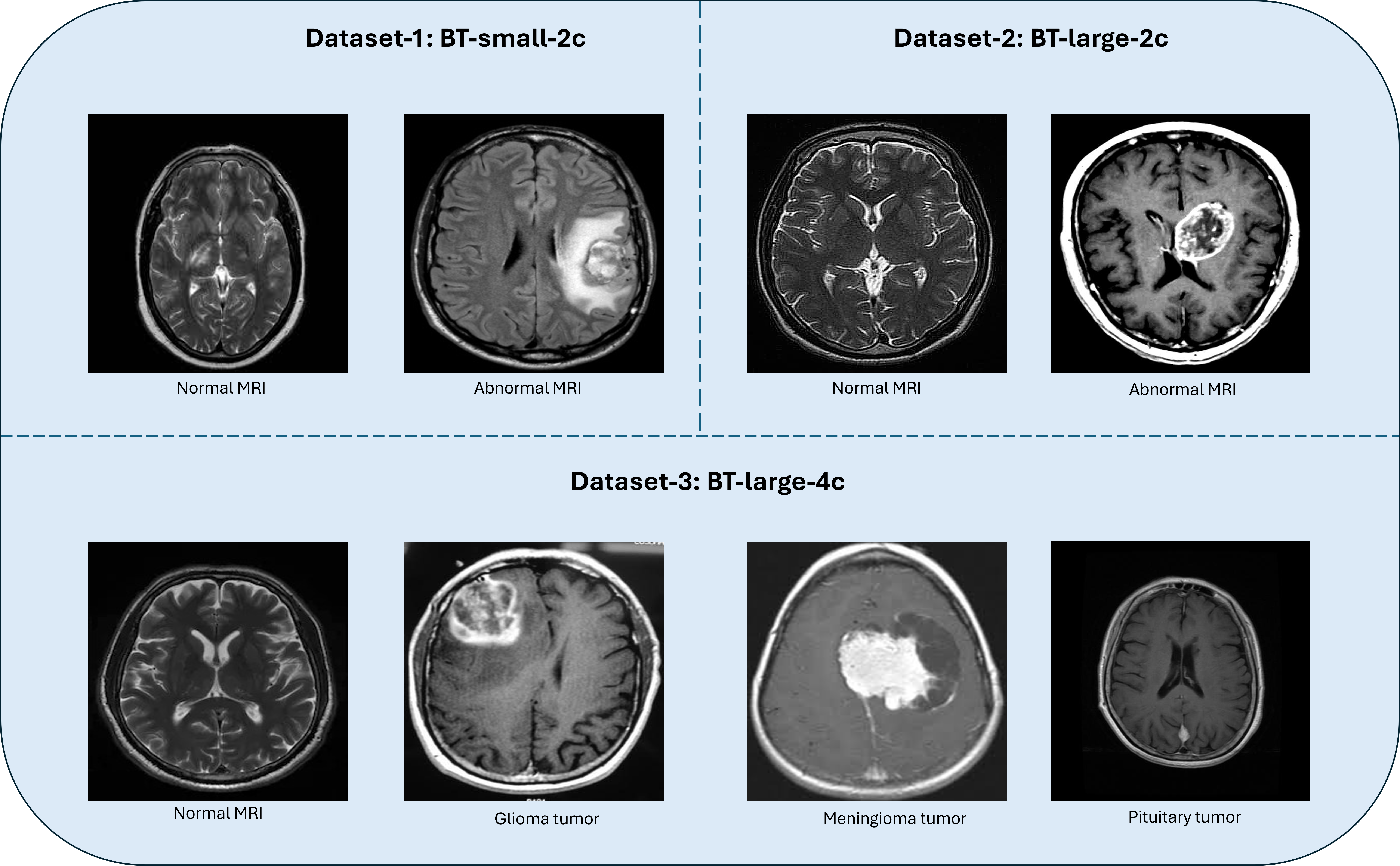}
     \caption{MRI Images with Corresponding Classes from Each Dataset. }
     \label{datasetimg}
 \end{figure*}

In this study, we conducted a series of experiments using three publicly available brain MRI datasets for brain tumor classification. The first dataset, referred to as BT-small-2c, was sourced from Kaggle \cite{chakrabarty2019brain} and contains 253 images, with 155 depicting tumors and 98 without tumors. The second dataset, named BT-large-2c, was also obtained from Kaggle and comprises 3,000 images evenly split between 1,500 tumor-containing and 1,500 non-tumor images \cite{Hamada2020}. The third dataset includes 3,064 T1-weighted images representing three distinct brain tumor types: gliomas, meningiomas, and pituitary tumors. We obtained this dataset also from the Kaggle website and labeled it BT-large-4c \cite{bhuvaji2020brain}. All these datasets provide a diverse foundation for evaluating brain tumor classification methodologies. Similarly, each dataset is split into a training set, comprising 80\% of the total data, and a test set, making up the remaining 20\%. Details of the datasets utilized in our experiments are presented in Table \ref{dataset}. Whereas, Figure \ref{datasetimg} displays sample brain MR images from the BT-small-2c, BT-large-2c, and BT-large-4c datasets, respectively.

\begin{table*}[!t]
\centering
\caption{illustrates each dataset details.}
\begin{tabular}{cccc}
\hline
\textbf{Types} & \textbf{Number of classes} & \textbf{Training set} & \textbf{Test set} \\ \hline
BT-small-2c    & 2                          & 202                   & 51                \\
BT-large-2c    & 2                          & 2400                  & 600               \\
BT-large-4c    & 4                          & 2611                  & 653               \\ \hline
\end{tabular}
\label{dataset}
\end{table*}

\subsection{Pre-Processing}
\label{pp}
\begin{figure*}[!ht]
     \centering
     \includegraphics[width=1\textwidth]{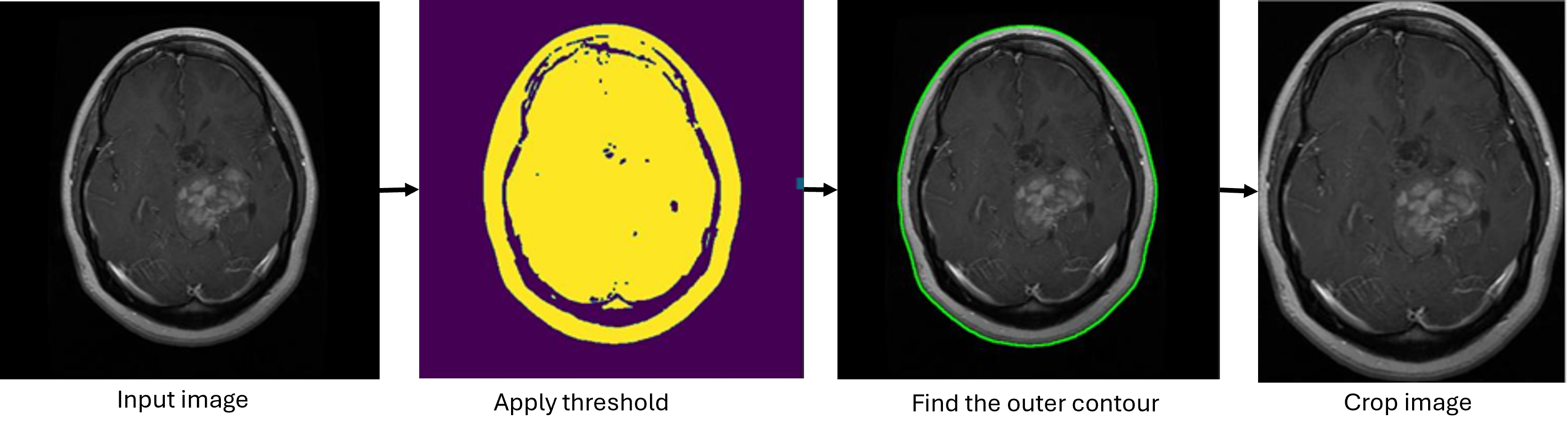}
     \caption{Pre-processing steps to crop the MRI images  }
     \label{preprocess}
 \end{figure*}

In our brain MRI datasets, most images include extraneous spaces and regions that hinder classification accuracy. To improve performance, it is essential to crop these images, removing irrelevant sections while preserving only the paramount data. We adopt the cropping method described in \cite{zhang2020finding}, which based on calculating extreme points. The cropping procedure for MR images, as depicted in Figure \ref{preprocess}, 
where thresholding is used to transform the images into a binary format, followed by dilation and erosion processes to minimize noise. Further, in the binary images, the largest contour is detected, and the 4 extreme points—topmost, bottommost, leftmost, and rightmost—are determined. Based on these two criteria, the images are cropped accordingly, and finally resized the cropped images with the help of bicubic interpolation. As bilinear interpolation create smoother curves and better handles the pronounced edge noise in MR images.

Additionally, we implemented image augmentation to address the small size of our MRI dataset. Image augmentation is a method that artificially expands a dataset by modifying the original images. This technique generates various versions of each image by tweaking factors like scale, rotation, position, brightness, and other properties. Research indicates \cite{perez2017effectiveness,yang2022image} that enhancing a dataset through augmentation can improve model classification accuracy more efficiently than gathering additional data. In the image augmentation process, we utilized two techniques—rotation and horizontal flipping—to generate extra training data. The rotation involved randomly turning the input image 90 degrees one or multiple times. Following this, horizontal flipping was applied to each rotated image to further expand the dataset.

\subsection{Deep Feature Extraction using ViTs and CNNs Models}
\label{feaExt}
\subsubsection{Vision Transformer}
\begin{figure*}[h]
     \includegraphics[width=1\textwidth]{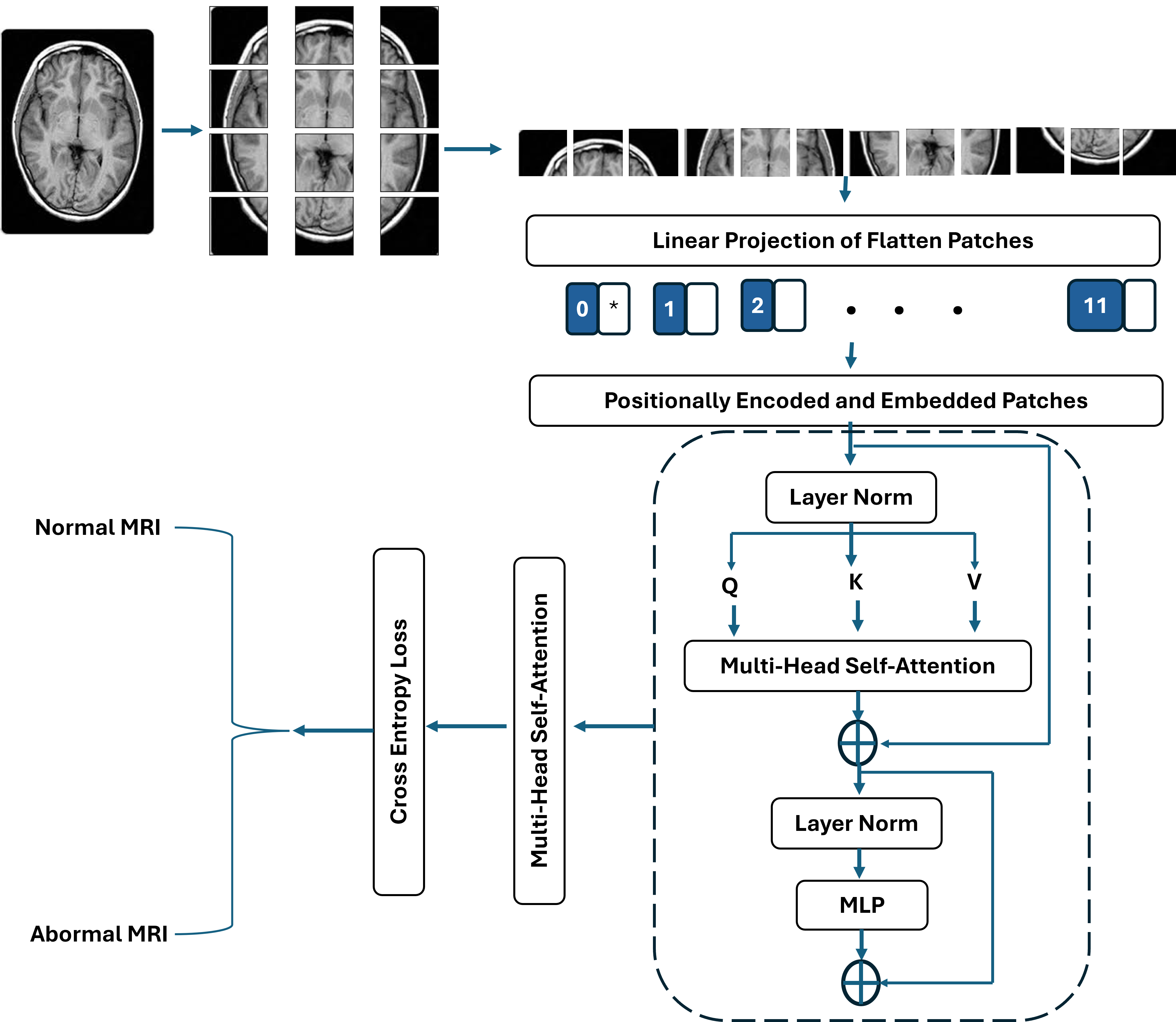}
     \caption{Overview of ViT Architecture. }
     \label{fig_vit_architecture}
 \end{figure*}

In June 2021, researchers at Google introduced a pure ViT architecture called ViT \cite{dosovitskiy2020image}, as shown in Fig. \ref{fig_vit_architecture}. This model demonstrated SOTA performance on extensive visual datasets like JFT-300M.  

The ViT model is designed for image classification by directly applying it to image patches. It adopts the transformer framework and is tailored primarily for processing 2D images. While standard transformers operate on 1D input sequences of token embeddings, ViT adapts this approach for 2D images by pre-processing the image. Specifically, an image $X$ with dimension $H \times W $ and $C$ channels is divided into a sequence of flattened 2D patches, denoted as $x_p$ with dimensions $N\times (P^2 \cdot C)$, Here, $P \times P$ represents the resolution of each patch, $C$ is the number of image channels, and $N$, the total number of patches, is computed as $N = (HW) / P^2$. 

The flattened patches are then mapped to a fixed-dimensional latent vector $d$, referred to as patch embeddings, using a linear projection. To enable classification, the model appends a special classification token $[cls]$ to the sequence of embeddings, functioning as the representation of the image. A classification head is attached to the embedding sequence $z_L^0$ during both the pre-training and fine-tuning stages. This classification head comprises a multi-layer perceptron (MLP) with a hidden layer during pre-training, while a single linear layer is used during fine-tuning. Additionally, to preserve positional information of pixels and patches, a 1D positional embedding matching the dimensions of the input embeddings is incorporated into the input sequence vector. Notably, ViT employs only the transformer's encoder component, incorporating layer normalization and an MLP head.

\begin{figure*}[!ht]
     \centering
     \includegraphics[width=1\textwidth]{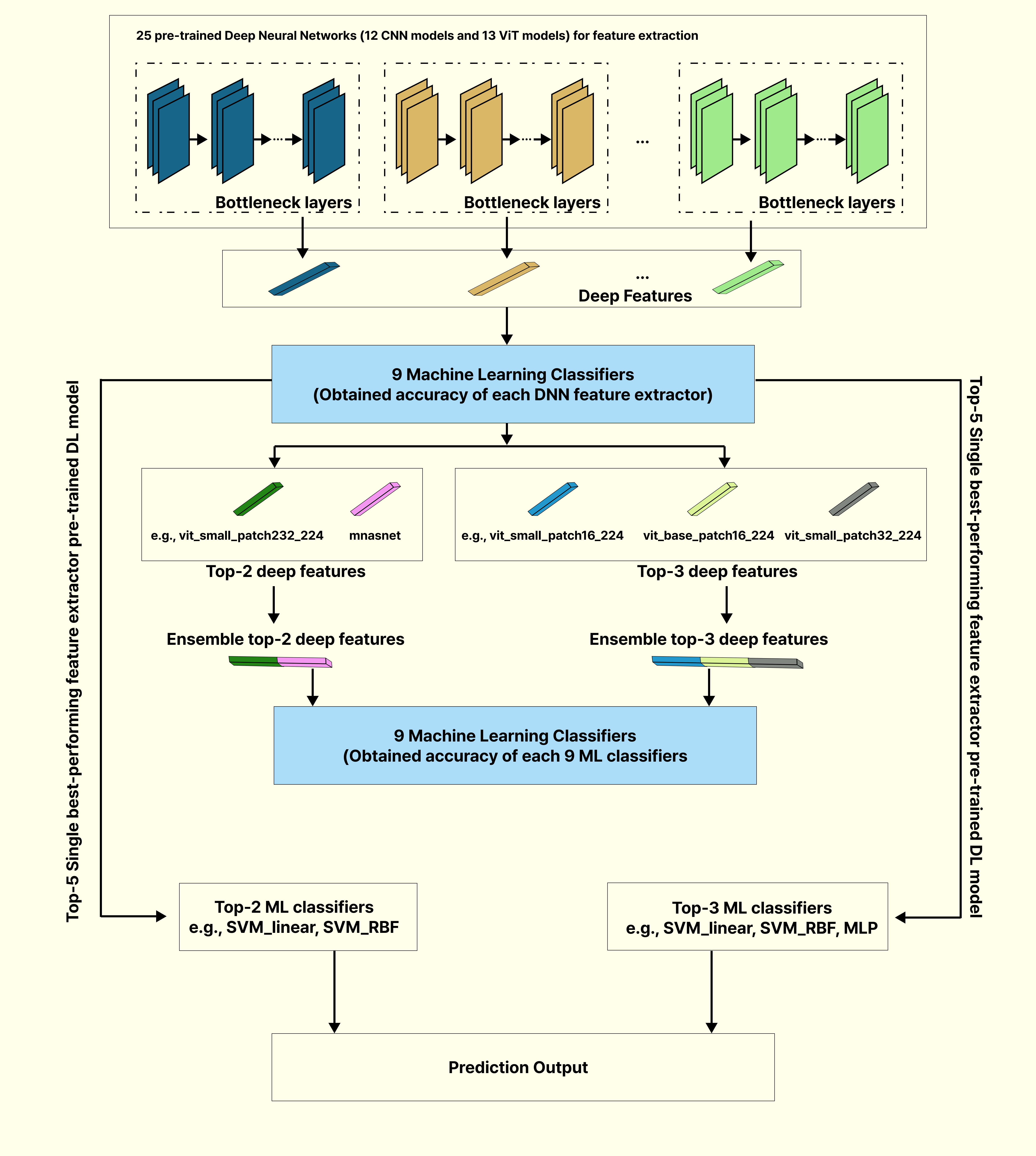}
     \caption{Our proposed double ensembling architecture for brain tumor classification. }
     \label{proposed}
 \end{figure*}

\textbf{Image Patching}: Instead of processing an image as a grid of pixels like CNNs, a ViTs divide the image into a sequence of fixed-size patches (e.g., 16\time16 pixels). Each patch is regarded as a ``token", analogous to a word in NLP.
\textbf{Flattening and Embedding}: These patches are flattened into vectors and passed through a linear layer to create patch embeddings, which are lower-dimensional representations of the patches.
\textbf{Positional Encoding}: Since Transformers lack a built-in sense of spatial order (unlike CNNs), positional encodings are incorporated into the patch embeddings to retain information about the patches' locations in the original image.
\textbf{Transformer Layers}: The sequence of patch embeddings goes through a series of Transformer encoder layers, comprising multi-head self-attention mechanisms and feed-forward networks. It allows the model to detect relationships between patches, regardless of their spatial distance.
\textbf{Classification Head}: For tasks like image classification, a special [CLS] token (inspired by BERT) is often added to the sequence. After processing, the output associated with this token is utilized to predict the class label via a simple feed-forward layer.

\textbf{Key Features}
\textbf{Self-Attention}: Unlike CNNs, which rely on local receptive fields, ViTs use self-attention to model global dependencies across the entire image, allowing them to ``see" relationships between distant patches.
\textbf{Scalability}: ViTs perform exceptionally well when trained on large datasets (e.g., ImageNet-21k or JFT-300M) and scale effectively with more data and compute.
\textbf{No Convolution}: ViTs abandon convolutions entirely, relying solely on the Transformer architecture.

\subsubsection{vision Networks}
\begin{figure*}[!t]
     \centering
     \includegraphics[width=1\textwidth]{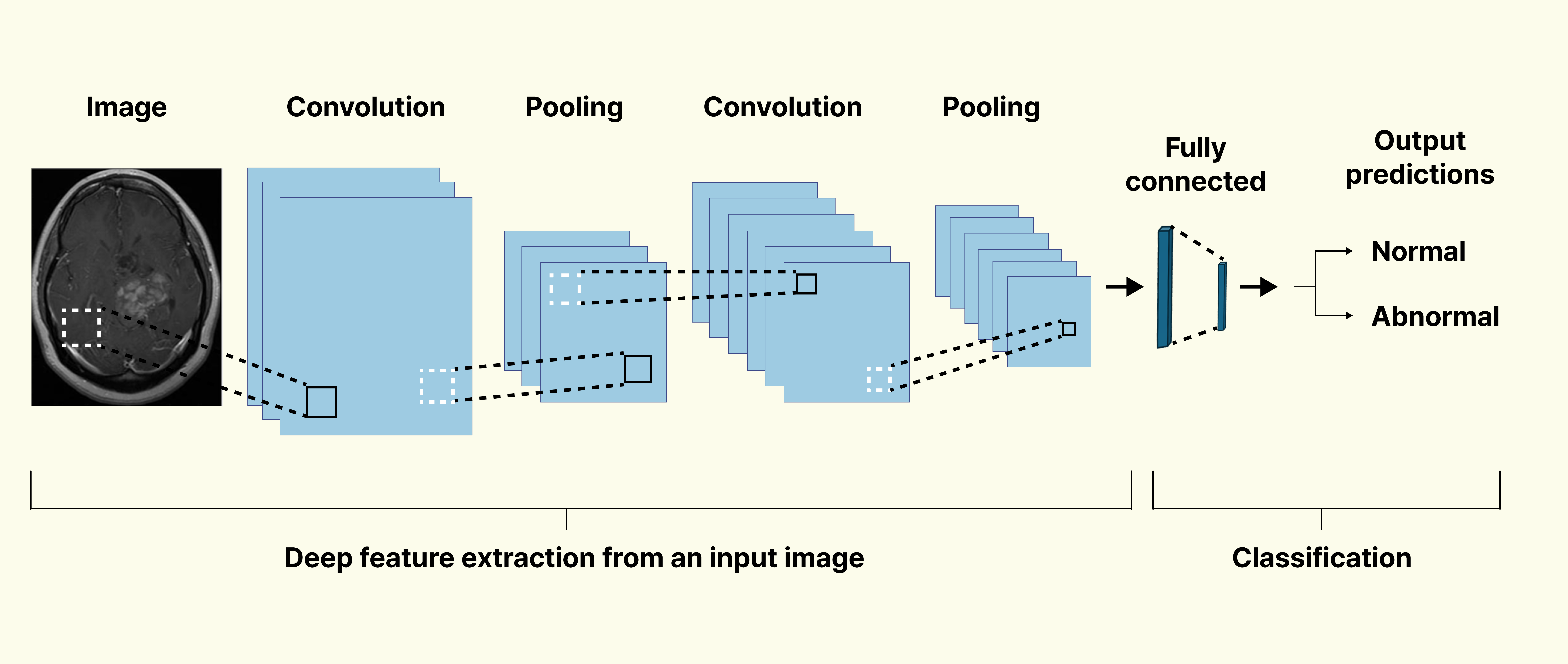}
     \caption{Convolutional neural network architecture. }
     \label{cnn}
 \end{figure*}

 CNNs use convolutional layers to extract meaningful features from input data. These layers apply filters to local regions of an image to compute the outputs of neurons, effectively capturing both spatial and temporal characteristics. Additionally, CNNs employ weight sharing in their convolutional layers, which helps to minimize the parameters in the model \cite{wu2017introduction,goyal2019learning}. As illustrated in Fig. \ref{cnn}, the CNNs are typically built from three key components. First, the convolutional layer is responsible for capturing spatial and temporal features from the input data. Second, a subsampling or max-pooling layer is used to downsample the input, reducing its dimensionality. Finally, the fully connected layer serves to classify the processed image into its respective categories as shown in Fig. \ref{cnn}.

\subsubsection{Transfer Learning}
In DL, transfer learning typically entails using a network pre-trained on a large dataset (such as ImageNet) and fine-tuning it for a smaller, domain-specific dataset such as BT-Small-2c-dataset. This method not only minimizes training time and data requirements but also improves model performance, particularly when annotated data is limited.

\begin{figure*}[!t]
     \centering
     \includegraphics[width=1\textwidth]{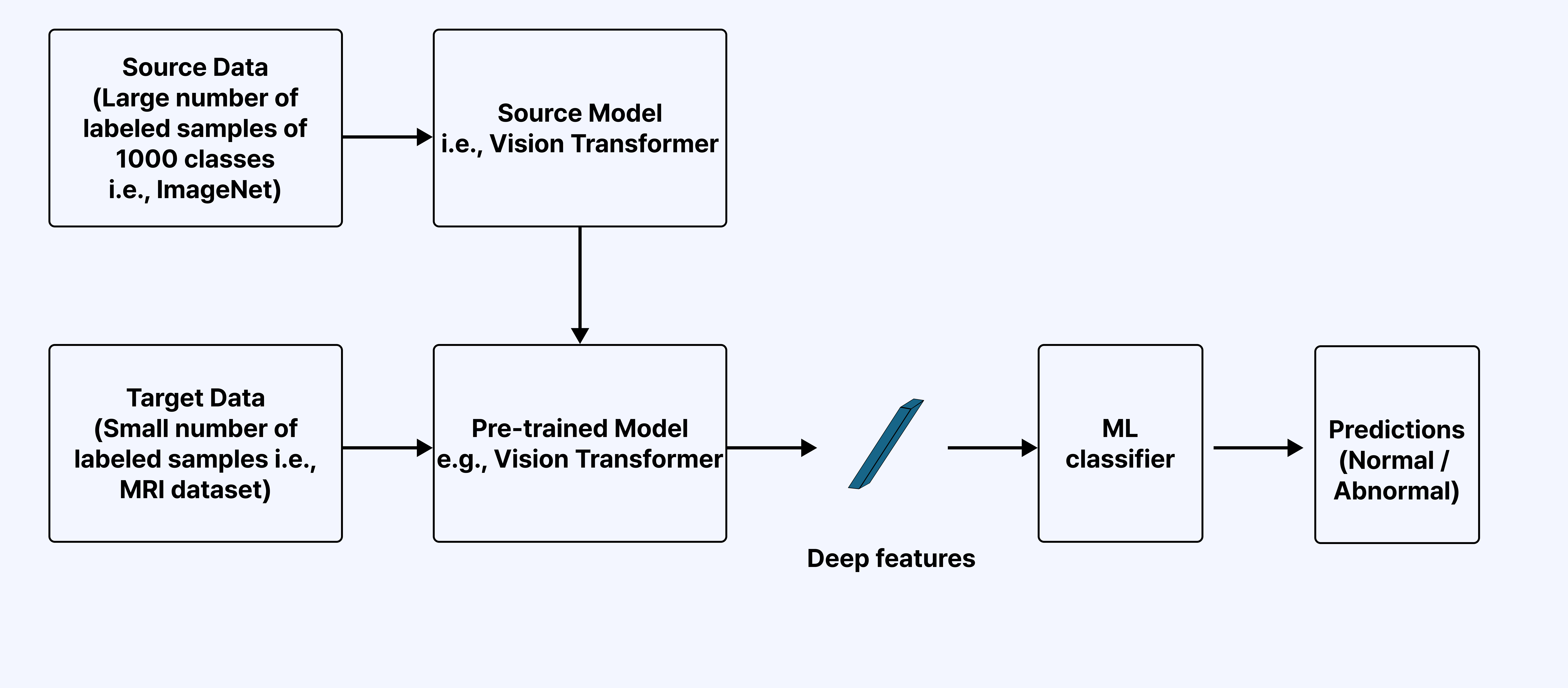}
     \caption{Methodology of transfer learning. }
     \label{tlearning}
 \end{figure*}

CNNs and ViTs typically achieve higher accuracy with large datasets, so when gathering a huge training data is impractical, transfer learning becomes a valuable alternative. As illustrated in Fig. \ref{tlearning}, where a model pre-trained on extensive benchmark datasets like ImageNet \cite{krizhevsky2012imagenet} can be repurposed as a feature extractor for tasks that use much smaller datasets, such as MRI scans. Recently, this technique has proven successful across various fields—including medical image classification and segmentation \cite{akccay2016transfer,baltruschat2019comparison,christodoulidis2016multisource} by significantly reducing both the training time required and the need for a large training dataset \cite{tajbakhsh2016convolutional,weiss2016survey}.

\subsubsection{Deep Feature Extraction}
\label{dpe}
We utilized both ViT-based and CNN-based models as DL feature extractors, leveraging their ability to autonomously identify and capture critical features without human intervention. Additionally, we adopted a transfer learning strategy to develop our feature extractor, given the small size of our MRI dataset. Training and fine-tuning a deep CNN, such as DenseNet-121, from scratch would be impractical in this context. Instead, we employed the pre-trained, fixed weights of each ViT and CNN model, originally trained on the extensive ImageNet dataset, to effectively extract deep features from brain MRI images. This approach enhances the robustness and efficiency of our feature extraction process. 

In this study, we utilized the following pre-trained ViT (vit\_base\_patch16\_224 , vit\_base\_patch32\_224, vit\_large\_patch16\_224, vit\_small\_patch32\_224, deit3\_small\_patch16\_224, vit\_base\_patch8\_224, vit\_tiny\_patch16\_224, vit\_small\_patch16\_224, vit\_base\_patch16\_384, vit\_tiny\_patch16\_384, vit\_small\_patch32\_384, vit\_small\_patch16\_384, vit\_base\_patch32\_384 \cite{dosovitskiy2020image, wu2020visual, deng2009imagenet}) and CNN (ResNet \cite{he2016deep}, DenseNet \cite{huang2017densely}, VGG \cite{simonyan2014very}, AlexNet \cite{krizhevsky2014one}, ResNext \cite{szegedy2016rethinking}, ShuffleNet \cite{xie2017aggregated}, MobileNet \cite{sandler2018mobilenetv2}, and MnasNet \cite{tan2019mnasnet}) models to extract the deep features from MRI as shown in Fig. \ref{cnn} and Fig. \ref{fig_vit_architecture}.

\subsection{Brain Tumor Classification using Machine Learning Classifiers} \label{MLcl}
Deep features extracted via pre-trained DL models are fed as inputs to various ML classifiers. In this paper, we have utilized numerous ML classifiers to categorize brain tumors. These classifiers include neural networks with an MLP, XGBoost, Gaussian Naïve Bayes (Gaussian NB), Adaptive Boosting (AdaBoost), k-NN, Random Forest (RF), Support Vector Machines (SVM) with three kernel types such as, linear, sigmoid, and radial basis function (RBF). We utilized the scikit-learn ML library \cite{pedregosa2011scikit} to implement these classifiers. The following subsections detail the ML classifiers and their hyper-parameter configurations applied in our brain tumor classification experiments.

\subsubsection{Multi layer Perceptron}

An MLP is a class of artificial neural networks composed of multiple layers of nodes, where each node (neuron) is connected to nodes in subsequent layers through directed edges. The primary goal of an MLP is to map input data to corresponding outputs by iteratively optimizing its parameters through backpropagation \cite{meng2021agwo}. \textbf{Forward Propagation}: The forward propagation process involves the computation of outputs at each layer based on the inputs and the weights and biases of the neurons. Mathematically, the output $h^{(l)}$ of the $l$-th layer is computed as:

\begin{equation}
    y^{(l)} = w^{(l)}h^{l-1)} + b^{(l)},
\end{equation}

\begin{equation}
    h^{(l)} = ReLU(z^{(l)}),
\end{equation}

where $w^{(l)}$ and $b^{(l)}$ represents the weights and biases of the $l-$th layer, $y^{(l)}$ denotes the linear transformation of the input, and $h^{(l-1)}$ is the output from the previous layer and ReLU is utilized as an activation function defined as follows:
\begin{equation}
    ReLU(x) = max(0, x)
\end{equation}
The optimization of the MLP is carried out using the Adam optimizer, an adaptive learning rate method that combines the benefits of momentum and RMSProp. The weights and biases are updated iteratively using the following rules:

\begin{equation}
M_t=\beta_1 M_{t-1}+\left(1-\beta_1\right) \nabla L_t, \\
\end{equation}
\begin{equation}
    V_t=\beta_2 V_{t-1}+\left(1-\beta_2\right)\left(\nabla L_t\right)^2, \\
\end{equation}

\begin{equation}
\hat{M}_t=\frac{M_t}{1-\beta_1^t}, \quad \hat{V}_t=\frac{V_t}{1-\beta_2^t}, \\
\end{equation}

\begin{equation}
  \theta_{t+1}=\theta_t-\eta \frac{\hat{M}_t}{\sqrt{\hat{V}_t}+\epsilon},  
\end{equation}

where $M_t$ and $V_t$ represent the first and second moment estimates, respectively, $\nabla {L_t}$  denotes the gradient at time $t$, $\eta$ is the learning rate (set to 0.001) and $\beta_1$, $\beta_2,$ and $\epsilon$ are  hyperparameters of the optimizer. 

\textbf{Learning Process}: The learning process of the MLP involves minimizing a loss function $\textbf{L}$, which quantifies the difference between predicted outputs $\hat{y}$ and actual outputs $y$. For a dataset with $N$ samples, the loss is typically computed as the mean squared error (MSE) for regression tasks:

\begin{equation}
    L=\frac{1}{N} \sum_{i=1}^N\left(y_i-\hat{y}_i\right)^2
\end{equation}

By iteratively adjusting the weights and biases through the Adam optimizer, the MLP gradually improves its ability to generalize and make accurate predictions. The combination of the ReLU activation function and the Adam solver ensures efficient training while mitigating issues such as vanishing gradients.

\subsubsection{Gaussian Naive Bayes}
This classifier is an ML model that assumes conditional independence among the features given the class label. In this study, we employ the Gaussian NB classifier as one of our ML approaches for classifying brain tumors. Using this classifier, the conditional probability \emph{P(y|X)} is computed by multiplying the individual conditional probabilities, leveraging the naïve assumption of independence between the attributes.

\begin{equation}
P(y|X) = \frac{P(y)P(X|y)}{P(X)} = \frac{P(y)
\prod_{i=1}^{n}P(x_i|y)}{P(X)}
\end{equation}

Here, {\emph{X}} represents a data instance derived from the deep features of a brain MR image, expressed as a feature vector \( (x_1, ..., x_n) \),. The variable {\emph{y}} denotes the target class—that is, the type of brain tumor—with two classes for the BT-small-2c and BT-large-2c MRI datasets, or four possible classes for the BT-large-4c dataset. Since \( P(X) \) is the same for all classes, classification of a given instance is based on the remaining terms as follows:


\begin{equation}
\hat{Y} = \arg\max_Y \, P(Y) \left( \prod_{i=1}^{n} P(x_i \mid Y) \right)
\end{equation}

where \( (x_i|Y) \) is computed under the assumption that the feature likelihood follows a Gaussian distribution, as expressed below:

\begin{equation}
p(x_i|Y) = \frac{1}{\sqrt{2\pi \sigma_Y^2}}exp( \frac{(x_i-\mu_Y)^2}{2\sigma_Y^2} )
\end{equation}
where the parameters \( \mu_Y \) and \( \sigma_Y \) are determined via maximum likelihood estimation.

Here, the smoothing parameter, which denotes the fraction of the highest variance among all features added to the variances to ensure computational stability, is configured at \( 10^{-9}\), aligning with the default setting in the scikit-learn ML library.

\subsubsection{AdaBoost}

AdaBoost, introduced in \cite{freund1997decision}, is an ensemble learning algorithm that improves overall performance by combining multiple classifiers. It creates a strong classifier through an iterative process that assigns weights to individual weak classifiers and adjusts them during each boosting iteration. The idea is to train on weighted data samples so that the combined model can accurately predict the class label—for instance, distinguishing between class datasets (i.e., BT-small-2c and BT-large-2c) or among four classes in the BT-large-4c dataset. Since any ML classifier that supports sample weights can serve as a base model, we selected the decision tree classifier due to its common use in AdaBoost. Additionally, we set the number of estimators to 100.

\subsubsection{K-Nearest Neighbors}
The k-NN algorithm is a straightforward classification method, where predictions are made directly from the stored training data. For example, when classifying a new instance (such as a deep feature extracted from a brain MR image), k-NN identifies the k training samples that are closest to the new instance by computing their distances. It then assigns the new instance the label that is most common among these k neighbors. This method works for binary classification as well as multi-class scenarios. While both Manhattan and Euclidean distances are frequently used to measure similarity, our approach employs the Euclidean distance metric. The Euclidean distance, \textit{d}, between two points \textit{x} and \textit{y} is calculated as follows:

\begin{equation}
d(X,Y) = \sqrt{ ( \sum\nolimits_{i=1}^{N}(X_i-Y_i)^2) }  
\end{equation}


\begin{equation}
    P = (Y_i),i,...,n
\end{equation}
where $Y_i$ represents a specific training sample in the training dataset, $n$ denotes the over all training samples, and $c_i$ corresponds to the class label associated with $Y_i$. 

\begin{itemize}
    \item During the testing phase, the distances between the new feature vector and the stored feature vectors from the training data are calculated. The new example is then classified based on the majority vote of its $k$-nearest neighbors.
\end{itemize}

The accuracy of the algorithm is evaluated during the testing phase using the correct classifications. If the performance is unsatisfactory, the value of $k$ can be adjusted to achieve a more acceptable level of accuracy. In this study, the neighbors size is varied from 1 to 4, and the value that resulted in the highest accuracy was selected.

\subsubsection{Random Forest}
RF, introduced by Breiman \cite{breiman2001random}, is an ensemble learning algorithm that constructs multiple decision trees using the bagging method. It classifies new data instances, such as deep features extracted from brain MRI images, into specific target classes. For the MRI datasets BT-small-2c and BT-large-2c, RF distinguishes between two classes, while for the BT-large-4c dataset, it identifies four classes.

During tree construction, RF randomly selects $n$ features to determine the best split point using the Gini index as the cost function. This random feature selection reduces correlation among the trees, thereby lowering the overall error rate of the ensemble. For prediction, the new data instance is passed through all the decision trees in the RF, with each tree providing a class prediction. The final class label for the data instance is decided by maximum voting, with the class receiving the most votes being designated as the predicted label.

In the study, the square root of the total number of features was used to determine the optimal split. The size of trees was tested from 1 to 150, and the configuration yielding the highest accuracy was chosen.

\subsubsection{Support Vector Machine}
SVM \cite{cortes1995support}, introduced by Vladimir N. Vapnik, are robust classification algorithms that utilize kernel functions to transform input data into higher-dimensional feature spaces, facilitating the determination of optimal separating hyperplanes.

\begin{equation}
f(x_i) =  \sum_{n=1}^{N}\alpha_n y_n K(x_n, x_i)+b 
\end{equation}

Here, the support vectors, denoted as \(x_n\), represent the deep features extracted from brain MR images. The Lagrange multipliers, \(\alpha_n\), are coefficients assigned to each support vector during the optimization process. The target classes, \(y_n\), correspond to the classification labels in the datasets used in this study: two binary-class datasets (normal and abnormal) and one dataset with four classes, where \(n=1,2,3,..., N\).

We employed the most widely utilized kernel functions within the SVM algorithm: (1) the linear kernel, (2) the sigmoid kernel, and (3) the RBF kernel as outlined in Table \ref{svms}. Additionally, the SVM relies on two critical hyperparameters: C and Gamma. The C hyperparameter governs the soft margin cost function, determining the impact of each support vector, while Gamma influences the degree of curvature in the decision boundary. We tested Gamma values of `scale' and 'auto' with values [0.1, 1, and 10], and C values of [0.1, 1, 10, 100], ultimately choosing the Gamma and C combination that yielded the highest accuracy.

\begin{table}[!ht]
\centering
	\caption{Kernel types and their required parameters.}
	\label{tab:2}
	\setlength{\tabcolsep}{5.6mm}
	\begin{tabular}{cccc}
		\toprule
		\textbf{Kernel} & \textbf{Equation} & \textbf{Parameters } \\
		\midrule
		Linear & \(  K(x_n,x_i) = (x_n,x_i) \) & - \\
		Sigmoid & \( K(x_n,x_i) = tanh (\gamma (x_n,x_i)+C) \) & \( \gamma, C \)   \\
		RBF & \( K(x_n,x_i) = exp(-\gamma\left\|x_n-x_i\right\|^2+C) \) & \( \gamma, C \)  \\
		\bottomrule
	\end{tabular}
    \label{svms}
\end{table}

\subsection{Hyperparamter tuning on machine learning models}
Hyperparameter optimization (HPO) involves identifying the most effective set of hyperparameter values and the optimal arrangement for categorical hyperparameters. This process aims to improve model performance by reducing a predefined loss function, leading to more accurate results with fewer mistakes. Hyperparameter tuning \cite{yu2020hyper} refers to the method of crafting the perfect model structure with an ideal hyperparameter setup. Since each ML algorithm comes with its own unique hyperparameters, manually adjusting them demands a thorough knowledge of the models and their corresponding hyperparameter settings. Some proposed \cite{tran2020hyper} automated hyperparameter tuning techniques, such as random search, grid search, and Bayesian optimization, offer greater adaptability compared to conventional approaches for choosing the best hyperparameters, ultimately boosting model performance more efficiently.

ML tasks can be described as developing a model M that reduces a predefined loss function $L(X_{Ts}; M)$ on a specific test set $X_{Ts}$, where the loss function (L) represents the error rate. A learning algorithm A utilizes a training set $X_{Tr}$ to construct the model M, frequently addressing a nonconvex optimization challenge. This learning algorithm A incorporates certain hyperparameters lambda $\lambda$, and the model M is defined as $M = A(X_{Tr})$; $\lambda$. The goal of HPO is to identify the optimal settings $\lambda^{ast}$ that produce an ideal model $M^{ast}$, which minimizes the loss function $L(X_{Ts}; M)$.


\begin{align}
\lambda^* &= \underset{\lambda}{\operatorname{argmin}} \, L(X_{Ts}; A(X_{Tr}; \lambda)) &= \underset{\lambda}{\operatorname{argmin}} \, F(X_{Ts}, X_{Tr}, A, \lambda, L)
\end{align}

Here, F represents the model’s objective function, which takes $\lambda$—a set of hyperparameters—and returns the associated loss. The loss function $L$ and the learning algorithm are selected, and the datasets $X_{Ts}$ (test set) and $X_{Tr}$ (training set) are provided \cite{claesen2015hyperparameter}. These elements vary based on the chosen model, the hyperparameter search space, and the selected ML classifiers.

In this study, hyperparameter tuning was uniformly applied to nine ML models, with each model undergoing the same optimization process using grid search. This approach aimed to determine the optimal hyperparameters for each model, striking a balance between accuracy and computational efficiency. By maintaining consistency in the tuning process, the models were evaluated under comparable conditions, ensuring a fair assessment of their predictive performance and generalization capabilities.

As illustrated in Table \ref{MLparamtuning}, to enhance the effectiveness of our ML models, we perform HPO through grid search methods \cite{belete2022grid}. The hyperparameter settings that produce the best outcomes on the validation set are chosen for each model. Table \ref{BTs2call}, \ref{BTlarge2c}, \ref{BTlarge4c_hypertune} depict the results obtained from our ML models following this hyperparameter tuning process.

\begin{table*}[!ht]
\centering
\caption{Selected hyperparameter with search space for ML classifiers.}
\scalebox{0.6}{
\begin{tabular}{cccc}
\hline
\textbf{Model} & \textbf{Hyperparameter}                                                                                                                                                            & \textbf{Search Space}                                                                                                                                                                                                                                 & \textbf{Type}                                                                                                    \\ \hline
XGBoost        & \begin{tabular}[c]{@{}c@{}}max\_depth\\ learning\_rate\\ subsample\\ n\_estimators\end{tabular}                                                                                    & \begin{tabular}[c]{@{}c@{}}{[}3, 5, 7{]},\\ {[}0.1, 0.01, 0.001{]}\\ {[}0.5, 0.7, 1{]}\\ {[}100, 200, 300{]}\end{tabular}                                                                                                                             & \begin{tabular}[c]{@{}c@{}}Discrete\\ Continuous\\ Continuous\\ Discrete\end{tabular}                            \\ \hline
MLP            & \begin{tabular}[c]{@{}c@{}}hidden\_layer\_sizes\\ activation\\ solver\\ max\_iter\\ momentum\end{tabular}                                                                          & \begin{tabular}[c]{@{}c@{}}{[}(50,), (100,22), (100,100, 50), (100, 50, 36, 30), (100, 100, 200, 150, 100){]}\\ {[}relu, tanh, logistic{]}\\ {[}adam, sgd, lbfgs{]}\\ {[}1000{]}\\ {[}0.9, 0.95, 0.99{]}\end{tabular}                                 & \begin{tabular}[c]{@{}c@{}}Discrete\\ \\ Categorical\\ Discrete\\ Continuous\end{tabular}                        \\ \hline
Gaussian NB    & \begin{tabular}[c]{@{}c@{}}var\_smoothing\\ priors\end{tabular}                                                                                                                    & \begin{tabular}[c]{@{}c@{}}{[}1e-9, 1e-8, 1e-7, 1e-6, 1e-5{]}\\ {[}None, {[}0.3, 0.7{]}, {[}0.4, 0.6{]}, {[}0.5, 0.5{]}{]}\end{tabular}                                                                                                               & \begin{tabular}[c]{@{}c@{}}Continuous\\ Continuous\end{tabular}                                                  \\ \hline
Adaboost       & \begin{tabular}[c]{@{}c@{}}n\_estimators\\ learning\_rate\end{tabular}                                                                                                             & \begin{tabular}[c]{@{}c@{}}{[}50, 70, 90, 120, 180, 200{]}\\ {[}0.001, 0.01, 0.1, 1, 10{]}\end{tabular}                                                                                                                                               &                                                                                                                  \\ \hline
KNN            & \begin{tabular}[c]{@{}c@{}}n\_neighbors\\ weights\\ algorithm\\ leaf\_size\\ p\\ metric\\ n\_jobs\end{tabular}                                                                     & \begin{tabular}[c]{@{}c@{}}list(range(1, 31))\\ {[}uniform, distance{]}\\ {[}autom ball\_tree, kd\_tree, brute{]}\\ list(range(10, 51, 5))\\ {[}1, 2{]}\\ {[}euclidean, manhattan, minkowski{]}\\ {[}-1{]}\end{tabular}                               & \begin{tabular}[c]{@{}c@{}}Discrete\\ Categorical\\ \\ Discrete\\ Discrete\\ Categorical\\ Discrete\end{tabular} \\ \hline
RF             & \begin{tabular}[c]{@{}c@{}}n\_estimators\\ max\_depth\\ min\_samples\_split\\ min\_samples\_leaf\\ max\_features\\ bootstrap\\ criterion\\ oob\_score\\ random\_state\end{tabular} & \begin{tabular}[c]{@{}c@{}}{[}100, 200, 300, 400, 500{]}\\ {[}None, 10, 20, 30, 40, 50{]}\\ {[}2, 5, 10{]}\\ {[}1, 2, 4{]}\\ {[}auto, sqrt, log2{]}\\ {[}True, False{]}\\ {[}gini, entropy{]}\\ {[}True, False{]}\\ {[}42{]}\end{tabular}             & \begin{tabular}[c]{@{}c@{}}Discrete\\ Discrete\\ Discrete\\ Discrete\\ Categorical\end{tabular}                  \\ \hline
SVM\_linear    & \begin{tabular}[c]{@{}c@{}}C\\ kernel\\ tol\\ class\_weight\\ random\_state\end{tabular}                                                                                           & \begin{tabular}[c]{@{}c@{}}{[}0.1, 1, 10, 100, 1000{]}\\ {[}linear{]}\\ {[}1e-3, 1e-4, 1e-5{]}\\ {[}None, balanced{]}\\ {[}42{]}\end{tabular}                                                                                                         & \begin{tabular}[c]{@{}c@{}}Continuous\\ Categorical\\ \\ \\ Discrete\end{tabular}                                \\ \hline
SVM\_sigmoid   & \begin{tabular}[c]{@{}c@{}}kernel\\ C\\ gamma\\ coef0\\ tol\\ class\_weight\\ shrinking\\ probability\\ cache\_size\\ random\_state\end{tabular}                                   & \begin{tabular}[c]{@{}c@{}}sigmoid\\ {[}0.1, 1. 10, 100{]}\\ {[}scale, auto{]}\\ {[}0.0, 0.1, 0.5, 1.0{]}\\ {[}1e-3, 1e-4, 1e-5{]}\\ {[}None, balanced{]}\\ {[}True, False{]}\\ {[}True, False{]}\\ {[}200.0, 500.0, 100.0{]}\\ {[}42{]}\end{tabular} & \begin{tabular}[c]{@{}c@{}}Continuous\\ Categorical\\ Continuous\end{tabular}                                    \\ \hline
SVM\_RBF       & \begin{tabular}[c]{@{}c@{}}C\\ gamma\\ kernel\\ class\_weight\\ shrinking\\ probability\\ tol\\ cache\_size\\ max\_iter\end{tabular}                                               & \begin{tabular}[c]{@{}c@{}}{[}0.1, 1, 10, 100{]}\\ {[}scale, auto, 0.1, 1, 10{]}\\ {[}rbf{]}\\ {[}None, balanced{]}\\ {[}True, False{]}\\ {[}True, False{]}\\ {[}1e-3, 1e-4{]}\\ {[}200, 500, 1000{]}\\ {[}-1, 1000, 5000{]}\end{tabular}             & \begin{tabular}[c]{@{}c@{}}Discrete\\ Categorical\\ \\ \\ \\ \\ \\ Discrete\end{tabular}                         \\ \hline
\end{tabular}
}
\label{MLparamtuning}
\end{table*}

\begin{table}[!hbt]
\centering
\caption{Accuracies of pre-trained CNN models and ViT models using fine-tune hyperparameter of ML classifiers on BT-small-2c dataset. The top-3 deep features were represented using \(\star \).}
\scalebox{0.5}{
\begin{tabular}{ccccccccccc}
\hline
\multicolumn{1}{l}{\textbf{\begin{tabular}[c]{@{}l@{}}Deep Feature from the \\ Pre-Trained CNN and                                                                                                                                                                       \\ ViT Model\end{tabular}}} & \multicolumn{10}{c}{\textbf{Machine Learning Classifier Accuracy}}                                                                                                                                        \\ \cline{2-11} 
\textbf{}                                                                                                                                                                                                                                                                                            & \textbf{XGBoost} & \textbf{MLP}    & \textbf{GaussianNB} & \textbf{Adaboost} & \textbf{KNN} & \textbf{RFClassifier} & \textbf{SVM\_linear} & \textbf{SVM\_sigmoid} & \textbf{SVM\_RBF} & \textbf{Average} \\ \hline
resnet50                                                                                                                                                                                                                                                                                             & 0.7371           & 0.7282          & 0.5976              & 0.6               & 0.7387       & 0.6677                & 0.7282               & 0.421                 & 0.8177            & 0.6707           \\
resnet101                                                                                                                                                                                                                                                                                            & 0.6444           & 0.6782          & 0.7871              & 0.625             & 0.7403       & 0.6516                & 0.6621               & 0.5137                & 0.6855            & 0.6653           \\
densenet121                                                                                                                                                                                                                                                                                          & 0.8089           & 0.8677          & 0.7726              & 0.75              & 0.7387       & 0.825                 & 0.8677               & 0.8516                & 0.8427            & 0.8139           \\
densenet169                                                                                                                                                                                                                                                                                          & 0.875            & 0.8677          & 0.8121              & 0.775             & 0.8032       & 0.85                  & 0.8355               & 0.7944                & 0.8589            & 0.8302           \\
vgg16                                                                                                                                                                                                                                                                                                & 0.9589           & 0.9089          & 0.7355              & 0.85              & 0.8048       & 0.875                 & 0.8516               & 0.9089                & 0.8927            & 0.8651           \\
vgg19                                                                                                                                                                                                                                                                                                & 0.8516           & 0.8605          & 0.7694              & 0.8589            & 0.846        & 0.7766                & 0.8355               & 0.8516                & 0.8855            & 0.8373           \\
alexnet                                                                                                                                                                                                                                                                                              & 0.8677           & 0.925           & 0.746               & 0.85              & 0.7726       & 0.8839                & 0.9339               & 0.7589                & 0.95              & 0.8542           \\
resnext50\_32x4d                                                                                                                                                                                                                                                                                     & 0.6371           & 0.7944          & 0.6315              & 0.625             & 0.6065       & 0.6339                & 0.7532               & 0.5                   & 0.7516            & 0.6592           \\
resnext101\_32x8d                                                                                                                                                                                                                                                                                    & 0.7839           & 0.8516          & 0.7226              & 0.625             & 0.7298       & 0.7427                & 0.8266               & 0.6532                & 0.8339            & 0.7522           \\
shufflenet\_v2\_x1\_0                                                                                                                                                                                                                                                                                & 0.7855           & 0.9089          & 0.6589              & 0.725             & 0.7637       & 0.7516                & 0.9089               & 0.9                   & 0.9               & 0.8114           \\
mobilenet\_v2                                                                                                                                                                                                                                                                                        & 0.7089           & 0.8427          & 0.6089              & 0.6589            & 0.6427       & 0.6839                & 0.8177               & 0.7839                & 0.8               & 0.7275           \\
mnasnet0\_5                                                                                                                                                                                                                                                                                          & 0.7605           & 0.8927          & 0.625               & 0.7339            & 0.6944       & 0.8089                & 0.8589               & 0.8589                & 0.7089            & 0.7713           \\
vit\_base\_patch16\_224 \(\star \)                                                                                                                                                                                                                                                                             & 1                & 0.975           & 0.8944              & 0.925             & 0.9105       & 0.925                 & 0.975                & 0.9339                & 0.975             & \textbf{0.946}   \\
vit\_base\_patch32\_224                                                                                                                                                                                                                                                                              & 0.9339           & 0.95            & 0.846               & 0.9089            & 0.9355       & 0.9589                & 0.9016               & 0.9339                & 0.9339            & 0.9225           \\
vit\_large\_patch16\_224                                                                                                                                                                                                                                                                             & 0.9              & 0.95            & 0.846               & 0.8089            & 0.8194       & 0.9                   & 0.975                & 0.975                 & 0.975             & 0.9055           \\
vit\_small\_patch32\_224                                                                                                                                                                                                                                                                             & 0.9427           & 0.95            & 0.9105              & 0.9589            & 0.9105       & 0.9339                & 0.95                 & 0.9589                & 0.975             & 0.9434           \\
deit3\_small\_patch16\_224                                                                                                                                                                                                                                                                           & 0.8589           & 0.9589          & 0.7032              & 0.8               & 0.8371       & 0.85                  & 0.9427               & 0.7427                & 0.9016            & 0.8439           \\
vit\_base\_patch8\_224                                                                                                                                                                                                                                                                               & 0.925            & 0.95            & 0.8782              & 0.925             & 0.8855       & 0.9                   & 0.9339               & 0.9339                & 0.95              & 0.9202           \\
vit\_tiny\_patch16\_224                                                                                                                                                                                                                                                                              & 0.975            & 0.975           & 0.8944              & 0.95              & 0.9032       & 0.925                 & 0.8516               & 0.9427                & 0.975             & 0.9324           \\
vit\_small\_patch16\_224 \(\star \)                                                                                                                                                                                                                                                                            & 1                & 0.975           & 0.8282              & 0.975             & 0.9016       & 0.95                  & 0.9339               & 0.975                 & 0.975             & \textbf{0.9460}  \\
vit\_base\_patch16\_384                                                                                                                                                                                                                                                                              & 0.9339           & 0.9589          & 0.9032              & 0.975             & 0.8855       & 0.95                  & 0.975                & 0.9339                & 0.9589            & 0.9416           \\
vit\_tiny\_patch16\_384                                                                                                                                                                                                                                                                              & 0.8927           & 0.8927          & 0.8032              & 0.95              & 0.8121       & 0.925                 & 0.8194               & 0.9339                & 0.9589            & 0.8875           \\
vit\_small\_patch32\_384                                                                                                                                                                                                                                                                             & 0.8677           & 0.9589          & 0.9089              & 0.925             & 0.9427       & 0.9339                & 0.9355               & 0.9339                & 0.9266            & 0.9259           \\
vit\_small\_patch16\_384                                                                                                                                                                                                                                                                             & 0.95             & 0.9339          & 0.8766              & 0.925             & 0.8121       & 0.9177                & 0.9177               & 0.9589                & 0.975             & 0.9185           \\
vit\_base\_patch32\_384                                                                                                                                                                                                                                                                              & 0.9              & 0.975           & 0.8137              & 0.925             & 0.9016       & 0.925                 & 0.975                & 0.9339                & 0.975             & 0.9249           \\ \hline
Average                                                                                                                                                                                                                                                                                              & 0.86             & \textbf{0.9012} & 0.7829              & 0.8251            & 0.8135       & 0.8458                & 0.8786               & 0.8355                & 0.8953            &                  \\ \hline
\end{tabular}
}
\label{BTs2call}
\end{table}

\begin{table}[!ht]
\centering
\caption{Accuracies of pre-trained CNN models and ViT models using fine tune hyperparameter of ML classifiers on BT-large-2c dataset. The top-3 deep features were represented using \(\star \).}
\scalebox{0.5}{
\begin{tabular}{ccccccccccc}
\hline
\multicolumn{1}{l}{\textbf{\begin{tabular}[c]{@{}l@{}}Deep Feature from the \\ Pre-Trained CNN and                                                                                                                                                                       \\ ViT Model\end{tabular}}} & \multicolumn{10}{c}{\textbf{Machine Learning Classifier Accuracy}}                                                                                                                                     \\ \cline{2-11} 
\textbf{}                                                                                                                                                                                                                                                                                            & \textbf{XGBoost} & \textbf{MLP} & \textbf{GaussianNB} & \textbf{Adaboost} & \textbf{KNN} & \textbf{RFClassifier} & \textbf{SVM\_linear} & \textbf{SVM\_sigmoid} & \textbf{SVM\_RBF} & \textbf{Average} \\ \hline
resnet50                                                                                                                                                                                                                                                                                             & 0.7667           & 0.8250       & 0.7850              & 0.7583            & 0.7967       & 0.7267                & 0.8417               & 0.5183                & 0.8800            & 0.7665           \\
resnet101                                                                                                                                                                                                                                                                                            & 0.7850           & 0.8367       & 0.7050              & 0.7750            & 0.8483       & 0.7183                & 0.8300               & 0.5417                & 0.8733            & 0.7681           \\
densenet121                                                                                                                                                                                                                                                                                          & 0.9383           & 0.9700       & 0.8500              & 0.9750            & 0.9817       & 0.9417                & 0.9600               & 0.9533                & 0.9733            & 0.9493           \\
densenet169                                                                                                                                                                                                                                                                                          & 0.9550           & 0.9700       & 0.8183              & 0.9750            & 0.9800       & 0.9383                & 0.9683               & 0.9550                & 0.9783            & 0.9487           \\
vgg16                                                                                                                                                                                                                                                                                                & 0.9283           & 0.9733       & 0.8300              & 0.9700            & 0.9633       & 0.9333                & 0.9700               & 0.9133                & 0.9817            & 0.9404           \\
vgg19                                                                                                                                                                                                                                                                                                & 0.9183           & 0.9733       & 0.7367              & 0.9683            & 0.9617       & 0.9383                & 0.9700               & 0.8333                & 0.9800            & 0.9200           \\
alexnet                                                                                                                                                                                                                                                                                              & 0.9267           & 0.9750       & 0.8250              & 0.9750            & 0.9783       & 0.9583                & 0.9633               & 0.6183                & 0.9833            & 0.9115           \\
resnext50\_32x4d                                                                                                                                                                                                                                                                                     & 0.7483           & 0.8167       & 0.5900              & 0.7383            & 0.7717       & 0.7067                & 0.8133               & 0.4833                & 0.8383            & 0.7230           \\
resnext101\_32x8d                                                                                                                                                                                                                                                                                    & 0.8083           & 0.8583       & 0.6500              & 0.8267            & 0.9133       & 0.7617                & 0.8633               & 0.6517                & 0.9067            & 0.8044           \\
shufflenet\_v2\_x1\_0                                                                                                                                                                                                                                                                                & 0.8717           & 0.9650       & 0.7200              & 0.9317            & 0.9767       & 0.8817                & 0.9650               & 0.9483                & 0.9733            & 0.9148           \\
mobilenet\_v2                                                                                                                                                                                                                                                                                        & 0.7800           & 0.9133       & 0.7100              & 0.8150            & 0.9217       & 0.7850                & 0.9050               & 0.8983                & 0.9067            & 0.8483           \\
mnasnet0\_5                                                                                                                                                                                                                                                                                          & 0.8433           & 0.9700       & 0.7033              & 0.9133            & 0.9433       & 0.9050                & 0.9650               & 0.9450                & 0.9633            & 0.9057           \\
vit\_base\_patch16\_224                                                                                                                                                                                                                                                                              & 0.9733           & 0.9933       & 0.8650              & 0.9817            & 0.9867       & 0.9783                & 0.9917               & 0.9383                & 0.9900            & 0.9665           \\
vit\_base\_patch32\_224                                                                                                                                                                                                                                                                              & 0.9717           & 0.9917       & 0.8767              & 0.9900            & 0.9883       & 0.9800                & 0.9917               & 0.9633                & 0.9950            & 0.9720           \\
vit\_large\_patch16\_224 \(\star \)                                                                                                                                                                                                                                                                             & 0.9850           & 0.9967       & 0.8683              & 0.9917            & 0.9850       & 0.9833                & 0.9967               & 0.9833                & 0.9967            & \textbf{0.9763}  \\
vit\_small\_patch32\_224                                                                                                                                                                                                                                                                             & 0.9700           & 0.9850       & 0.8917              & 0.9900            & 0.9933       & 0.9683                & 0.9717               & 0.9367                & 0.9950            & 0.9669           \\
deit3\_small\_patch16\_224                                                                                                                                                                                                                                                                           & 0.9167           & 0.9917       & 0.7983              & 0.9733            & 0.9700       & 0.9517                & 0.9550               & 0.8550                & 0.9900            & 0.9335           \\
vit\_base\_patch8\_224                                                                                                                                                                                                                                                                               & 0.9683           & 0.9933       & 0.8550              & 0.9850            & 0.9850       & 0.9733                & 0.9950               & 0.8817                & 0.9933            & 0.9589           \\
vit\_tiny\_patch16\_224                                                                                                                                                                                                                                                                              & 0.9533           & 0.9800       & 0.8650              & 0.9850            & 0.9850       & 0.9700                & 0.9533               & 0.8950                & 0.9867            & 0.9526           \\
vit\_small\_patch16\_224                                                                                                                                                                                                                                                                             & 0.9700           & 0.9917       & 0.8583              & 0.9900            & 0.9850       & 0.9800                & 0.9750               & 0.9383                & 0.9933            & 0.9646           \\
vit\_base\_patch16\_384                                                                                                                                                                                                                                                                              & 0.9750           & 0.9850       & 0.8950              & 0.9883            & 0.9867       & 0.9767                & 0.9850               & 0.9633                & 0.9850            & 0.9711           \\
vit\_tiny\_patch16\_384                                                                                                                                                                                                                                                                              & 0.9650           & 0.9933       & 0.8083              & 0.9883            & 0.9883       & 0.9833                & 0.9733               & 0.9000                & 0.9933            & 0.9548           \\
vit\_small\_patch32\_384                                                                                                                                                                                                                                                                             & 0.9767           & 0.9917       & 0.9067              & 0.9950            & 0.9900       & 0.9883                & 0.9750               & 0.9483                & 0.9967            & 0.9743           \\
vit\_small\_patch16\_384                                                                                                                                                                                                                                                                             & 0.9767           & 0.9917       & 0.8333              & 0.9900            & 0.9900       & 0.9900                & 0.9750               & 0.9450                & 0.9917            & 0.9648           \\
vit\_base\_patch32\_384 \(\star \)                                                                                                                                                                                                                                                                              & 0.9817           & 0.9917       & 0.8833              & 0.9933            & 0.9850       & 0.9883                & 0.9900               & 0.9650                & 0.9950            & \textbf{0.9748}           \\ \hline
Average                                                                                                                                                                                                                                                                                              & 0.9141           & 0.9569       & 0.8051              & 0.9385            & 0.9542       & 0.9163                & 0.9497               & 0.8549                & \textbf{0.9656}   &                  \\ \hline
\end{tabular}
}
\label{BTlarge2c}
\end{table}

\begin{table}[]
\centering
\caption{Accuracies of pre-trained CNN models and ViT models with fine tune hyperparameter ML classifiers on BT-large-4c dataset. The top-3 deep features were represented using \(\star \).}
\scalebox{0.5}{
\begin{tabular}{ccccccccccc}
\hline
\multicolumn{1}{l}{\textbf{\begin{tabular}[c]{@{}l@{}}Deep Feature from the \\ Pre-Trained CNN and                                                                                                                                                                       \\ ViT Model\end{tabular}}} & \multicolumn{10}{c}{\textbf{Machine Learning Classifier Accuracy}}                                                                                                                                        \\ \cline{2-11} 
\textbf{}                                                                                                                                                                                                                                                                                            & \textbf{XGBoost} & \textbf{MLP} & \textbf{GaussianNB} & \textbf{Adaboost} & \textbf{KNN}    & \textbf{RFClassifier} & \textbf{SVM\_linear} & \textbf{SVM\_sigmoid} & \textbf{SVM\_RBF} & \textbf{Average} \\ \hline
resnet50                                                                                                                                                                                                                                                                                             & 0.5965           & 0.7825       & 0.4749              & 0.7625            & 0.7057          & 0.7617                & 0.7853               & 0.2405                & 0.7927            & 0.6558           \\
resnet101                                                                                                                                                                                                                                                                                            & 0.5939           & 0.7271       & 0.3372              & 0.7118            & 0.7385          & 0.7669                & 0.7360               & 0.2743                & 0.7213            & 0.6230           \\
densenet121                                                                                                                                                                                                                                                                                          & 0.6896           & 0.7268       & 0.4551              & 0.8125            & 0.7769          & 0.7546                & 0.7316               & 0.6462                & 0.7254            & 0.7021           \\
densenet169                                                                                                                                                                                                                                                                                          & 0.7157           & 0.7123       & 0.4262              & 0.7687            & 0.7399          & 0.7387                & 0.7165               & 0.6530                & 0.7094            & 0.6867           \\
vgg16   \(\star \)                                                                                                                                                                                                                                                                                             & 0.6540           & 0.7491       & 0.5991              & 0.7606            & 0.7490          & 0.7424                & 0.7742               & 0.6232                & 0.7727            & 0.7138           \\
vgg19                                                                                                                                                                                                                                                                                                & 0.6529           & 0.7500       & 0.5452              & 0.7463            & 0.7219          & 0.7366                & 0.7575               & 0.5519                & 0.7766            & 0.6932           \\
alexnet                                                                                                                                                                                                                                                                                              & 0.6817           & 0.7332       & 0.5025              & 0.7249            & 0.7727          & 0.7391                & 0.7711               & 0.4606                & 0.7663            & 0.6836           \\
resnext50\_32x4d                                                                                                                                                                                                                                                                                     & 0.5975           & 0.6324       & 0.4111              & 0.7161            & 0.7330          & 0.7459                & 0.6624               & 0.2339                & 0.6993            & 0.6035           \\
resnext101\_32x8d                                                                                                                                                                                                                                                                                    & 0.5518           & 0.7184       & 0.3969              & 0.7151            & 0.7458          & 0.7359                & 0.6710               & 0.2278                & 0.7416            & 0.6116           \\
shufflenet\_v2\_x1\_0                                                                                                                                                                                                                                                                                & 0.6232           & 0.7372       & 0.5629              & 0.7111            & 0.7866          & 0.7810                & 0.7536               & 0.6803                & 0.7381            & 0.7082           \\
mobilenet\_v2                                                                                                                                                                                                                                                                                        & 0.5934           & 0.7234       & 0.5707              & 0.7064            & 0.7031          & 0.7161                & 0.7241               & 0.6652                & 0.7330            & 0.6817           \\
mnasnet0\_5 \(\star \)                                                                                                                                                                                                                                                                                         & 0.6114           & 0.7484       & 0.6710              & 0.7058            & 0.7304          & 0.7710                & 0.7454               & 0.7047                & 0.7450            & 0.7148           \\
vit\_base\_patch16\_224                                                                                                                                                                                                                                                                              & 0.6714           & 0.8020       & 0.4940              & 0.7263            & 0.7638          & 0.7213                & 0.7522               & 0.6134                & 0.7717            & 0.7018           \\
vit\_base\_patch32\_224                                                                                                                                                                                                                                                                              & 0.6835           & 0.7572       & 0.4860              & 0.7437            & 0.7922          & 0.7460                & 0.7404               & 0.6303                & 0.7375            & 0.7019           \\
vit\_large\_patch16\_224                                                                                                                                                                                                                                                                             & 0.6752           & 0.7419       & 0.4496              & 0.7111            & 0.7751          & 0.7188                & 0.7404               & 0.5986                & 0.7085            & 0.6799           \\
vit\_small\_patch32\_224  \(\star \)                                                                                                                                                                                                                                                                           & 0.7167           & 0.7690       & 0.5493              & 0.7767            & 0.8065          & 0.7852                & 0.7608               & 0.5995                & 0.7795            & \textbf{0.7270}  \\
deit3\_small\_patch16\_224                                                                                                                                                                                                                                                                           & 0.6675           & 0.7391       & 0.5040              & 0.7400            & 0.7592          & 0.7847                & 0.7120               & 0.5672                & 0.7404            & 0.6905           \\
vit\_base\_patch8\_224                                                                                                                                                                                                                                                                               & 0.6624           & 0.7699       & 0.4917              & 0.7274            & 0.7333          & 0.7210                & 0.7648               & 0.5666                & 0.7455            & 0.6870           \\
vit\_tiny\_patch16\_224                                                                                                                                                                                                                                                                              & 0.6746           & 0.7926       & 0.4537              & 0.7329            & 0.7829          & 0.7373                & 0.7088               & 0.5518                & 0.7564            & 0.6879           \\
vit\_small\_patch16\_224                                                                                                                                                                                                                                                                             & 0.6697           & 0.7795       & 0.5317              & 0.7425            & 0.8183          & 0.7619                & 0.7360               & 0.5843                & 0.7560            & 0.7089           \\
vit\_base\_patch16\_384                                                                                                                                                                                                                                                                              & 0.6917           & 0.7399       & 0.4930              & 0.7494            & 0.7819          & 0.7748                & 0.7419               & 0.6177                & 0.7492            & 0.7044           \\
vit\_tiny\_patch16\_384                                                                                                                                                                                                                                                                              & 0.6375           & 0.7706       & 0.4344              & 0.7143            & 0.7864          & 0.7113                & 0.6912               & 0.5184                & 0.7442            & 0.6676           \\
vit\_small\_patch32\_384                                                                                                                                                                                                                                                                             & 0.6420           & 0.7723       & 0.4858              & 0.7197            & 0.7804          & 0.7310                & 0.7451               & 0.5888                & 0.7772            & 0.6936           \\
vit\_small\_patch16\_384                                                                                                                                                                                                                                                                             & 0.6330           & 0.7676       & 0.5083              & 0.7667            & 0.7809          & 0.7579                & 0.7481               & 0.6260                & 0.7489            & 0.7042           \\
vit\_base\_patch32\_384                                                                                                                                                                                                                                                                              & 0.6751           & 0.7699       & 0.4963              & 0.7692            & 0.7535          & 0.7599                & 0.7773               & 0.6445                & 0.7497            & 0.7106           \\ \hline
Average                                                                                                                                                                                                                                                                                              & 0.6505           & 0.7485       & 0.4932              & 0.7385            & \textbf{0.7607} & 0.7480                & 0.7379               & 0.5468                & 0.7474            &                  \\ \hline
\end{tabular}
}
\label{BTlarge4c_hypertune}
\end{table}

\section{Experimental setup}
\label{experimental}
The experimental setup details are described in their corresponding subsections.

\subsection{Implementation details}
In this work, we utilized 13 pre-trained ViT-based models and 12 pre-trained CNN-based models as a feature extractor, as mentioned in section \ref{dpe}. We freeze the weights of the bottleneck layers in these deep neural network models that were pre-trained on the ImageNet \cite{krizhevsky2012imagenet} dataset. Additionally, we employ nine distinct ML classifiers as mentioned in section \ref{MLcl}. Prior to training, we preprocess the input images as outlined in Section \ref{pp}. We conducted all the experiments on an RTX 3090.

\subsubsection{Deep Feature Evaluation and Selection}
In this work, we have evaluated the extracted deep features such as 12 pre-trained CNN models and 13 pre-trained ViT models using different ML classifiers as mentioned in section \ref{MLcl} and select the top three deep features for each of the three MRI datasets based on the average accuracy achieved across nine different ML classifiers. If two or more models achieve identical accuracy, the model with the lowest standard deviation is preferred, as it indicates more consistent performance across different data subsets. Additionally, when dealing with similar pre-trained models (e.g., resnet50 and resnet101), if both rank among the top performers, the model with the lower accuracy is excluded in favor of the next best-performing model to ensure diversity and potentially enhance generalization. The main reason for considering this approach is that deep features derived from two similar models occupy overlapping feature spaces, leading to redundancy and reduced diversity in the ensemble of these features. The top three deep features are then utilized in our ensemble module, which is detailed in the subsequent subsection.

\subsubsection{Ensemble of Deep Features}
Ensemble learning seeks to enhance performance and mitigate the risks associated with relying on a single, potentially underperforming feature from one model by merging multiple features from various models into a unified predictive feature. This technique can be categorized into feature ensemble and classifier ensemble based on the level of integration. Feature ensemble combines feature sets that are subsequently input into a classifier to produce the final result, whereas classifier ensemble merges the outputs from multiple classifiers, using voting techniques to determine the ultimate outcome. Given that the feature set provides more comprehensive details about MR images compared to the output set from individual classifiers, integrating at the feature level is anticipated to yield superior classification outcomes. In this study, we adopt a feature ensemble 
as well as classifier ensemble approach. In our feature ensemble approach, we integrate deep features extracted from top-2 as well as top-3 distinct pre-trained models (either CNNs or ViTs) by concatenating them into a single feature sequence. For example, if vit\_base\_patch\_16\_224, vit\_small\_patch\_32\_224, and vit\_small\_patch\_16\_224 are identified as the top-performing models, their respective deep features are combined to form a unified feature vector. This concatenated feature set is then input into ML classifiers to predict the final output. Additionally, we explore all possible pairwise combinations of deep features from these top three models, feeding each combination into the classifiers. This comparative analysis allows us to assess the effectiveness of the three-feature ensemble against two-feature combinations in our experiments.

\section{Results} \label{results}
The experimental results were derived from three distinct datasets for brain tumor classification tasks. The initial experiment is structured to evaluate the performance of various pre-trained models with different ML classifiers using default hyperparameters. Table \ref{BT-small-2c-default-param}, \ref{BTlarg2c_defaultpara}, and \ref{BTlarge4c_def-para} shows the results of the first experiments on BT-small-2c, BT-large-2c, and BT-large-4c dataset.

In the second experiment, we conduct hyperparameter tuning of each ML classifier as shown in Table \ref{MLparamtuning}. Table \ref{BTs2call}, \ref{BTlarge2c}, and \ref{BTlarge4c_hypertune} presents the results obtained from our ML models after conducting hyperparameter tuning. After hyperparameter tuning, the performance of each ML classifier is significantly improved in classifying the brain tumor as compare to the default hyperparameters of ML classifier. For instance, using BT-small-2c dataset, if we compare the results of ML classifiers default hyperparaters as shown in Table \ref{BT-small-2c-default-param} with tune hyperparameters of ML classifier as shown in Table \ref{MLparamtuning}, the ML classifiers with fine tune hyperparameter out-perform the ML classifier with default hyperparameters.

The third experiment aims to demonstrate the advantages of ensembling the top two or three deep features, identified from the first and second experiments, with various ML classifiers.  As illustrated in Table \ref{BT-small-2c-default-param}, the features from vit\_base\_patch\_16\_224, vit\_small\_patch\_32\_224, and vit\_small\_patch\_16\_224 are selected as top three deep features on BT-small-2c dataset. Whereas, Table \ref{BTlarg2c_defaultpara} shows vit\_base\_patch32\_224, vit\_large\_patch16\_224, and vit\_base\_patch32\_384 features are selected as the top three deep features on BT-large-2c dataset. In BT-large-4c dataset, the top three deep feature networks are vgg16, mnasnet0\_5, and vit\_small\_patch\_32\_224 as shown in Table \ref{BTlarge4c_hypertune}. 

The last experiment consists of ensembling the top 2 and top 3 ML classifiers using only the enhanced version (i.e., with the combination of normalization, PCA, and SMOTE) as depicted in Table \ref{ML-ensemble-2c-small}, \ref{ensemble_ML-BT-large2c}, and \ref{4c_ML_ensemble}. It can be observed from the results that the ensemble of ML classifiers provides the best results for brain tumor classification as shown in Table \ref{ML-ensemble-2c-small}, \ref{ensemble_ML-BT-large2c}, and \ref{4c_ML_ensemble}, by outperforming the other combination with a high margin.

\begin{table}[!ht]
\centering
\caption{Performance of pre-trained CNN models and ViT models using default parameter of ML classifiers on original BT-small-2c dataset. The top-3 deep features were represented using \(\star \).}
\scalebox{0.5}{
\begin{tabular}{ccccccccccc}
\hline
\multicolumn{1}{l}{\textbf{\begin{tabular}[c]{@{}l@{}}Deep Feature from the \\ Pre-Trained CNN and                                                                                                                                                                       \\ ViT Model\end{tabular}}} & \multicolumn{10}{c}{\textbf{Machine Learning Classifier Accuracy}}                                                                                                                                     \\ \cline{2-11} 
\textbf{}                                                                                                                                                                                                                                                                                            & \textbf{XGBoost} & \textbf{MLP} & \textbf{GaussianNB} & \textbf{Adaboost} & \textbf{KNN} & \textbf{RFClassifier} & \textbf{SVM\_linear} & \textbf{SVM\_sigmoid} & \textbf{SVM\_RBF} & \textbf{Average} \\ \hline
resnet50                                                                                                                                                                                                                                                                                             & 0.5782           & 0.7282       & 0.5976              & 0.5960            & 0.7153       & 0.6750                & 0.7282               & 0.4210                & 0.7427            & 0.6425           \\
resnet101                                                                                                                                                                                                                                                                                            & 0.6266           & 0.6694       & 0.7871              & 0.6621            & 0.6331       & 0.6000                & 0.6621               & 0.5137                & 0.6177            & 0.6413           \\
densenet121                                                                                                                                                                                                                                                                                          & 0.8589           & 0.5000       & 0.8371              & 0.8177            & 0.7565       & 0.6839                & 0.8677               & 0.8516                & 0.8589            & 0.7814           \\
densenet169                                                                                                                                                                                                                                                                                          & 0.8177           & 0.8444       & 0.8048              & 0.8177            & 0.7726       & 0.8500                & 0.8355               & 0.7944                & 0.8339            & 0.8190           \\
vgg16                                                                                                                                                                                                                                                                                                & 0.9089           & 0.8927       & 0.6589              & 0.8016            & 0.7226       & 0.8250                & 0.8516               & 0.9089                & 0.8927            & 0.8292           \\
vgg19                                                                                                                                                                                                                                                                                                & 0.7855           & 0.8605       & 0.6177              & 0.7460            & 0.7903       & 0.7516                & 0.8355               & 0.8516                & 0.8516            & 0.7878           \\
alexnet                                                                                                                                                                                                                                                                                              & 0.8927           & 0.9089       & 0.6121              & 0.8839            & 0.6597       & 0.7839                & 0.9339               & 0.7589                & 0.8677            & 0.8113           \\
resnext50\_32x4d                                                                                                                                                                                                                                                                                     & 0.6694           & 0.7855       & 0.6315              & 0.6121            & 0.6242       & 0.6677                & 0.7532               & 0.5000                & 0.7750            & 0.6687           \\
resnext101\_32x8d                                                                                                                                                                                                                                                                                    & 0.6944           & 0.8516       & 0.7226              & 0.7855            & 0.7048       & 0.6750                & 0.8266               & 0.6532                & 0.8750            & 0.7543           \\
shufflenet\_v2\_x1\_0                                                                                                                                                                                                                                                                                & 0.8266           & 0.9000       & 0.6339              & 0.7298            & 0.7637       & 0.7839                & 0.9089               & 0.9000                & 0.8750            & 0.8135           \\
mobilenet\_v2                                                                                                                                                                                                                                                                                        & 0.7339           & 0.6452       & 0.6250              & 0.6782            & 0.5992       & 0.5839                & 0.8177               & 0.7839                & 0.6000            & 0.6741           \\
mnasnet0\_5                                                                                                                                                                                                                                                                                          & 0.8516           & 0.7839       & 0.6000              & 0.6782            & 0.7016       & 0.7089                & 0.8589               & 0.8589                & 0.6839            & 0.7473           \\
vit\_base\_patch16\_224 \(\star \)                                                                                                                                                                                                                                                                              & 0.9750           & 0.9750       & 0.8944              & 0.9355            & 0.8371       & 0.9500                & 0.9750               & 0.9339                & 0.9750            & 0.9390           \\
vit\_base\_patch32\_224                                                                                                                                                                                                                                                                              & 0.9177           & 0.9339       & 0.8460              & 0.9589            & 0.9032       & 0.9589                & 0.9016               & 0.9339                & 0.9589            & 0.9237           \\
vit\_large\_patch16\_224                                                                                                                                                                                                                                                                             & 0.9500           & 0.8839       & 0.8460              & 0.8355            & 0.7766       & 0.9000                & 0.9750               & 0.9750                & 0.9500            & 0.8991           \\
vit\_small\_patch32\_224 \(\star \)                                                                                                                                                                                                                                                                            & 0.9427           & 0.9750       & 0.9105              & 0.9105            & 0.8944       & 0.9589                & 0.9500               & 0.9589                & 0.9750            & 0.9418           \\
deit3\_small\_patch16\_224                                                                                                                                                                                                                                                                           & 0.8589           & 0.8766       & 0.7032              & 0.8177            & 0.7565       & 0.8250                & 0.9427               & 0.7427                & 0.8927            & 0.8240           \\
vit\_base\_patch8\_224                                                                                                                                                                                                                                                                               & 0.8839           & 0.9500       & 0.8782              & 0.8677            & 0.9105       & 0.9250                & 0.9339               & 0.9339                & 0.9500            & 0.9148           \\
vit\_tiny\_patch16\_224                                                                                                                                                                                                                                                                              & 0.9250           & 0.9750       & 0.8944              & 0.8339            & 0.9032       & 0.9000                & 0.8516               & 0.9427                & 0.9500            & 0.9084           \\
vit\_small\_patch16\_224 \(\star \)                                                                                                                                                                                                                                                                            & 0.9500           & 0.9750       & 0.8282              & 0.9839            & 0.9355       & 0.9250                & 0.9339               & 0.9750                & 0.9750            & \textbf{0.9424}  \\
vit\_base\_patch16\_384                                                                                                                                                                                                                                                                              & 0.9589           & 0.9339       & 0.9032              & 0.9427            & 0.9177       & 0.9250                & 0.9750               & 0.9339                & 0.9339            & 0.9360           \\
vit\_tiny\_patch16\_384                                                                                                                                                                                                                                                                              & 0.8677           & 0.9339       & 0.8032              & 0.8266            & 0.8460       & 0.9250                & 0.8194               & 0.9339                & 0.9339            & 0.8766           \\
vit\_small\_patch32\_384                                                                                                                                                                                                                                                                             & 0.8589           & 0.9589       & 0.9089              & 0.7855            & 0.9105       & 0.9000                & 0.9355               & 0.9339                & 0.9177            & 0.9011           \\
vit\_small\_patch16\_384                                                                                                                                                                                                                                                                             & 0.9500           & 0.9750       & 0.8766              & 0.8605            & 0.9089       & 0.9250                & 0.9177               & 0.9589                & 1.0000            & 0.9303           \\
vit\_base\_patch32\_384                                                                                                                                                                                                                                                                              & 0.9250           & 0.9250       & 0.8137              & 0.8750            & 0.8694       & 0.9250                & 0.9750               & 0.9339                & 0.9339            & 0.9084           \\ \hline
Average                                                                                                                                                                                                                                                                                              & 0.8483           & 0.8656       & 0.7694              & 0.8097            & 0.7925       & 0.8213                & \textbf{0.8786}      & 0.8355                & 0.8728            &                  \\ \hline
\end{tabular}
}
\label{BT-small-2c-default-param}
\end{table}

\begin{table}[!ht]
\centering
\caption{Accuracies of pre-trained CNN models of BT-large-2c dataset without preprocessing using default parameters of ML classifiers. }
\scalebox{0.6}{
\begin{tabular}{cccccccccl}
\hline
\multicolumn{1}{l}{\textbf{\begin{tabular}[c]{@{}l@{}}Deep Feature from the \\ Pre-Trained CNN Model\end{tabular}}} & \multicolumn{9}{c}{\textbf{Machine Learning Classifier Accuracy}}                                                                                                                                                                                                                                                   \\ \cline{2-10} 
\textbf{}                                                                                                           & \textbf{MLP}                & \textbf{Gaussian NB}        & \textbf{Adaboost}           & \textbf{KNN}                & {\color[HTML]{67626E} \textbf{RF Classifier}} & {\color[HTML]{67626E} \textbf{SVM\_linear}} & {\color[HTML]{67626E} \textbf{SVM\_sigmoid}} & \textbf{SVM\_RBF}           & \textbf{Average} \\ \hline
{\color[HTML]{67626E} resnet50}                                                                                     & {\color[HTML]{67626E} 0.86} & {\color[HTML]{67626E} 0.76} & {\color[HTML]{67626E} 0.8}  & {\color[HTML]{67626E} 0.7}  & {\color[HTML]{67626E} 0.62}                   & {\color[HTML]{67626E} 0.82}                 & {\color[HTML]{67626E} 0.64}                  & {\color[HTML]{67626E} 0.82} & 0.75             \\
{\color[HTML]{67626E} resnet101}                                                                                    & {\color[HTML]{67626E} 0.86} & {\color[HTML]{67626E} 0.86} & {\color[HTML]{67626E} 0.76} & {\color[HTML]{67626E} 0.78} & {\color[HTML]{67626E} 0.68}                   & {\color[HTML]{67626E} 0.88}                 & {\color[HTML]{67626E} 0.66}                  & {\color[HTML]{67626E} 0.8}  & 0.79             \\
{\color[HTML]{67626E} densenet121}                                                                                  & {\color[HTML]{67626E} 0.9}  & {\color[HTML]{67626E} 0.86} & {\color[HTML]{67626E} 0.84} & {\color[HTML]{67626E} 0.88} & {\color[HTML]{67626E} 0.78}                   & {\color[HTML]{67626E} \textbf{0.9}}         & {\color[HTML]{67626E} 0.88}                  & {\color[HTML]{67626E} 0.86} & \textbf{0.9}     \\
{\color[HTML]{67626E} densenet169}                                                                                  & {\color[HTML]{67626E} 0.92} & {\color[HTML]{67626E} 0.86} & {\color[HTML]{67626E} 0.78} & {\color[HTML]{67626E} 0.84} & {\color[HTML]{67626E} 0.82}                   & {\color[HTML]{67626E} 0.88}                 & {\color[HTML]{67626E} 0.86}                  & {\color[HTML]{67626E} 0.86} & 0.85             \\
{\color[HTML]{67626E} vgg16}                                                                                        & {\color[HTML]{67626E} 0.86} & {\color[HTML]{67626E} 0.74} & {\color[HTML]{67626E} 0.86} & {\color[HTML]{67626E} 0.78} & {\color[HTML]{67626E} 0.74}                   & {\color[HTML]{67626E} 0.9}                  & {\color[HTML]{67626E} 0.88}                  & {\color[HTML]{67626E} 0.9}  & 0.83             \\
{\color[HTML]{67626E} vgg19}                                                                                        & {\color[HTML]{67626E} 0.86} & {\color[HTML]{67626E} 0.78} & {\color[HTML]{67626E} 0.84} & {\color[HTML]{67626E} 0.72} & {\color[HTML]{67626E} 0.76}                   & {\color[HTML]{67626E} 0.88}                 & {\color[HTML]{67626E} 0.84}                  & {\color[HTML]{67626E} 0.9}  & 0.82             \\
{\color[HTML]{67626E} alexnet}                                                                                      & {\color[HTML]{67626E} 0.92} & {\color[HTML]{67626E} 0.8}  & {\color[HTML]{67626E} 0.92} & {\color[HTML]{67626E} 0.84} & {\color[HTML]{67626E} 0.8}                    & {\color[HTML]{67626E} 0.9}                  & {\color[HTML]{67626E} 0.84}                  & {\color[HTML]{67626E} 0.9}  & 0.87             \\
{\color[HTML]{67626E} resnext50\_32x4d}                                                                             & {\color[HTML]{67626E} 0.82} & {\color[HTML]{67626E} 0.78} & {\color[HTML]{67626E} 0.76} & {\color[HTML]{67626E} 0.68} & {\color[HTML]{67626E} 0.64}                   & {\color[HTML]{67626E} 0.82}                 & {\color[HTML]{67626E} 0.62}                  & {\color[HTML]{67626E} 0.74} & 0.73             \\
{\color[HTML]{67626E} resnext101\_32x8d}                                                                            & {\color[HTML]{67626E} 0.8}  & {\color[HTML]{67626E} 0.78} & {\color[HTML]{67626E} 0.74} & {\color[HTML]{67626E} 0.84} & {\color[HTML]{67626E} 0.64}                   & {\color[HTML]{67626E} 0.78}                 & {\color[HTML]{67626E} 0.68}                  & {\color[HTML]{67626E} 0.84} & 0.8              \\
{\color[HTML]{67626E} shufflenet\_v2\_x1\_0}                                                                        & {\color[HTML]{67626E} 0.86} & {\color[HTML]{67626E} 0.7}  & {\color[HTML]{67626E} 0.8}  & {\color[HTML]{67626E} 0.82} & {\color[HTML]{67626E} 0.7}                    & {\color[HTML]{67626E} 0.84}                 & {\color[HTML]{67626E} 0.84}                  & {\color[HTML]{67626E} 0.78} & 0.79             \\
{\color[HTML]{67626E} mobilenet\_v2}                                                                                & {\color[HTML]{67626E} 0.82} & {\color[HTML]{67626E} 0.68} & {\color[HTML]{67626E} 0.72} & {\color[HTML]{67626E} 0.74} & {\color[HTML]{67626E} 0.66}                   & {\color[HTML]{67626E} 0.82}                 & {\color[HTML]{67626E} 0.8}                   & {\color[HTML]{67626E} 0.72} & 0.75             \\
{\color[HTML]{67626E} mnasnet0\_5}                                                                                  & {\color[HTML]{67626E} 0.84} & {\color[HTML]{67626E} 0.7}  & {\color[HTML]{67626E} 0.86} & {\color[HTML]{67626E} 0.84} & {\color[HTML]{67626E} 0.68}                   & {\color[HTML]{67626E} 0.88}                 & {\color[HTML]{67626E} 0.84}                  & {\color[HTML]{67626E} 0.86} & 0.81             \\ \hline
Average                                                                                                             & \textbf{0.86}               & 0.78                        & 0.8                         & 0.8                         & 0.71                                          & 0.86                                        & 0.78                                         & 0.83                        &                  \\ \hline
\end{tabular}
}
\label{nopreprocessing}
\end{table}

\begin{table}[!ht]
\centering
\caption{Accuracies of pre-trained CNN models and ViT models using default parameter of ML classifiers on BT-large-2c dataset. The top-3 deep features were represented using \(\star \).}
\scalebox{0.5}{
\begin{tabular}{ccccccccccc}
\hline
\multicolumn{1}{l}{\textbf{\begin{tabular}[c]{@{}l@{}}Deep Feature from the \\ Pre-Trained CNN and                                                                                                                                                                       \\ ViT Model\end{tabular}}} & \multicolumn{10}{c}{\textbf{Machine Learning Classifier Accuracy}}                                                                                                                                     \\ \cline{2-11} 
\textbf{}                                                                                                                                                                                                                                                                                            & \textbf{XGBoost} & \textbf{MLP} & \textbf{GaussianNB} & \textbf{Adaboost} & \textbf{KNN} & \textbf{RFClassifier} & \textbf{SVM\_linear} & \textbf{SVM\_sigmoid} & \textbf{SVM\_RBF} & \textbf{Average} \\ \hline
resnet50                                                                                                                                                                                                                                                                                             & 0.7400           & 0.8283       & 0.7850              & 0.7117            & 0.6967       & 0.7017                & 0.8417               & 0.5183                & 0.8667            & 0.7433           \\
resnet101                                                                                                                                                                                                                                                                                            & 0.8000           & 0.8267       & 0.7050              & 0.7250            & 0.7900       & 0.7567                & 0.8300               & 0.5417                & 0.8717            & 0.7607           \\
densenet121                                                                                                                                                                                                                                                                                          & 0.9633           & 0.9783       & 0.8383              & 0.8867            & 0.9700       & 0.9433                & 0.9600               & 0.9533                & 0.9650            & 0.9398           \\
densenet169                                                                                                                                                                                                                                                                                          & 0.9717           & 0.9683       & 0.8183              & 0.8933            & 0.9783       & 0.9567                & 0.9683               & 0.9550                & 0.9733            & 0.9426           \\
vgg16                                                                                                                                                                                                                                                                                                & 0.9517           & 0.9717       & 0.7717              & 0.8450            & 0.9267       & 0.9417                & 0.9700               & 0.9133                & 0.9767            & 0.9187           \\
vgg19                                                                                                                                                                                                                                                                                                & 0.9483           & 0.9717       & 0.6917              & 0.8350            & 0.9150       & 0.9350                & 0.9700               & 0.8333                & 0.9633            & 0.8959           \\
alexnet                                                                                                                                                                                                                                                                                              & 0.9533           & 0.9800       & 0.7800              & 0.8533            & 0.9417       & 0.9617                & 0.9633               & 0.6183                & 0.9717            & 0.8915           \\
resnext50\_32x4d                                                                                                                                                                                                                                                                                     & 0.7367           & 0.8150       & 0.5900              & 0.6783            & 0.7083       & 0.6783                & 0.8133               & 0.4833                & 0.8383            & 0.7046           \\
resnext101\_32x8d                                                                                                                                                                                                                                                                                    & 0.8317           & 0.8383       & 0.6500              & 0.7750            & 0.8950       & 0.8083                & 0.8633               & 0.6517                & 0.9083            & 0.8024           \\
shufflenet\_v2\_x1\_0                                                                                                                                                                                                                                                                                & 0.8917           & 0.9700       & 0.6917              & 0.7633            & 0.9617       & 0.8750                & 0.9650               & 0.9483                & 0.9683            & 0.8928           \\
mobilenet\_v2                                                                                                                                                                                                                                                                                        & 0.7717           & 0.9150       & 0.6717              & 0.6817            & 0.8550       & 0.7650                & 0.9050               & 0.8983                & 0.9017            & 0.8183           \\
mnasnet0\_5                                                                                                                                                                                                                                                                                          & 0.8767           & 0.9633       & 0.6650              & 0.7233            & 0.9300       & 0.8933                & 0.9650               & 0.9450                & 0.9517            & 0.8793           \\
vit\_base\_patch16\_224                                                                                                                                                                                                                                                                              & 0.9833           & 0.9900       & 0.8650              & 0.9300            & 0.9817       & 0.9817                & 0.9917               & 0.9383                & 0.9867            & 0.9609           \\
vit\_base\_patch32\_224                                                                                                                                                                                                                                                                              & 0.9833           & 0.9950       & 0.8767              & 0.9300            & 0.9850       & 0.9867                & 0.9917               & 0.9633                & 0.9900            & 0.9669           \\
vit\_large\_patch16\_224 \(\star \)                                                                                                                                                                                                                                                                            & 0.9917           & 0.9983       & 0.8683              & 0.9400            & 0.9833       & 0.9833                & 0.9967               & 0.9833                & 0.9933            & \textbf{0.9709}  \\
vit\_small\_patch32\_224                                                                                                                                                                                                                                                                             & 0.9817           & 0.9917       & 0.8917              & 0.9250            & 0.9950       & 0.9717                & 0.9717               & 0.9367                & 0.9833            & 0.9609           \\
deit3\_small\_patch16\_224                                                                                                                                                                                                                                                                           & 0.9683           & 0.9900       & 0.7983              & 0.8683            & 0.9633       & 0.9517                & 0.9550               & 0.8550                & 0.9650            & 0.9239           \\
vit\_base\_patch8\_224                                                                                                                                                                                                                                                                               & 0.9767           & 0.9933       & 0.8550              & 0.9167            & 0.9783       & 0.9783                & 0.9950               & 0.8817                & 0.9900            & 0.9517           \\
vit\_tiny\_patch16\_224                                                                                                                                                                                                                                                                              & 0.9783           & 0.9867       & 0.8650              & 0.8800            & 0.9833       & 0.9733                & 0.9533               & 0.8950                & 0.9750            & 0.9433           \\
vit\_small\_patch16\_224                                                                                                                                                                                                                                                                             & 0.9850           & 0.9933       & 0.8583              & 0.9083            & 0.9817       & 0.9800                & 0.9750               & 0.9383                & 0.9883            & 0.9565           \\
vit\_base\_patch16\_384                                                                                                                                                                                                                                                                              & 0.9783           & 0.9817       & 0.8950              & 0.9233            & 0.9817       & 0.9833                & 0.9850               & 0.9633                & 0.9817            & 0.9637           \\
vit\_tiny\_patch16\_384                                                                                                                                                                                                                                                                              & 0.9767           & 0.9900       & 0.8083              & 0.8917            & 0.9850       & 0.9750                & 0.9733               & 0.9000                & 0.9933            & 0.9437           \\
vit\_small\_patch32\_384 \(\star \)                                                                                                                                                                                                                                                                            & 0.9850           & 0.9917       & 0.9067              & 0.9150            & 0.9900       & 0.9833                & 0.9750               & 0.9483                & 0.9833            & 0.9643           \\
vit\_small\_patch16\_384                                                                                                                                                                                                                                                                             & 0.9867           & 0.9883       & 0.8333              & 0.9367            & 0.9883       & 0.9850                & 0.9750               & 0.9450                & 0.9883            & 0.9585           \\
vit\_base\_patch32\_384 \(\star \)                                                                                                                                                                                                                                                                             & 0.9883           & 0.9933       & 0.8833              & 0.9450            & 0.9867       & 0.9833                & 0.9900               & 0.9650                & 0.9967            & 0.9702           \\ \hline
Average                                                                                                                                                                                                                                                                                              & 0.9280           & 0.9564       & 0.7945              & 0.8513            & 0.9341       & 0.9173                & 0.9497               & 0.8549                & \textbf{0.9589}   &                  \\ \hline
\end{tabular}
}
\label{BTlarg2c_defaultpara}
\end{table}

\begin{table}[!ht]
\centering
\caption{Accuracies of pre-trained CNN models and ViT models using default hyperparameter of ML classifiers on BT-large-4c dataset. The top-3 deep features were represented using \(\star \). }
\scalebox{0.5}{
\begin{tabular}{ccccccccccc}
\hline
\multicolumn{1}{l}{\textbf{\begin{tabular}[c]{@{}l@{}}Deep Feature from the \\ Pre-Trained CNN and                                                                                                                                                                       \\ ViT Model\end{tabular}}} & \multicolumn{10}{c}{\textbf{Machine Learning Classifier Accuracy}}                                                                                                                                        \\ \cline{2-11} 
\textbf{}                                                                                                                                                                                                                                                                                            & \textbf{XGBoost} & \textbf{MLP}    & \textbf{GaussianNB} & \textbf{Adaboost} & \textbf{KNN} & \textbf{RFClassifier} & \textbf{SVM\_linear} & \textbf{SVM\_sigmoid} & \textbf{SVM\_RBF} & \textbf{Average} \\ \hline
resnet50                                                                                                                                                                                                                                                                                             & 0.7579           & 0.7666          & 0.4749              & 0.3755            & 0.5291       & 0.3782                & 0.7853               & 0.2405                & 0.7515            & 0.5621           \\
resnet101                                                                                                                                                                                                                                                                                            & 0.7347           & 0.7420          & 0.3372              & 0.4156            & 0.6232       & 0.3534                & 0.7360               & 0.2743                & 0.7020            & 0.5465           \\
densenet121                                                                                                                                                                                                                                                                                          & 0.7643           & 0.7266          & 0.4381              & 0.5412            & 0.6916       & 0.3755                & 0.7316               & 0.6462                & 0.7010            & 0.6240           \\
densenet169                                                                                                                                                                                                                                                                                          & 0.7510           & 0.7223          & 0.4299              & 0.4700            & 0.6434       & 0.3856                & 0.7165               & 0.6530                & 0.6935            & 0.6072           \\
vgg16                                                                                                                                                                                                                                                                                                & 0.7703           & 0.7478          & 0.6040              & 0.5225            & 0.5836       & 0.3566                & 0.7742               & 0.6232                & 0.7215            & 0.6337           \\
vgg19                                                                                                                                                                                                                                                                                                & 0.7592           & 0.7539          & 0.5299              & 0.4696            & 0.5875       & 0.3043                & 0.7575               & 0.5519                & 0.7113            & 0.6028           \\
alexnet                                                                                                                                                                                                                                                                                              & 0.7382           & 0.7407          & 0.5335              & 0.4642            & 0.6371       & 0.3547                & 0.7711               & 0.4606                & 0.6698            & 0.5967           \\
resnext50\_32x4d                                                                                                                                                                                                                                                                                     & 0.7303           & 0.6885          & 0.4111              & 0.3767            & 0.5235       & 0.3479                & 0.6624               & 0.2339                & 0.6436            & 0.5131           \\
resnext101\_32x8d                                                                                                                                                                                                                                                                                    & 0.7344           & 0.7200          & 0.3969              & 0.3945            & 0.6297       & 0.3424                & 0.6710               & 0.2278                & 0.7138            & 0.5367           \\
shufflenet\_v2\_x1\_0                                                                                                                                                                                                                                                                                & 0.7344           & 0.7561          & 0.5464              & 0.4086            & 0.6641       & 0.3852                & 0.7536               & 0.6803                & 0.7285            & 0.6286           \\
mobilenet\_v2                                                                                                                                                                                                                                                                                        & 0.6991           & 0.7200          & 0.5508              & 0.3599            & 0.4743       & 0.3366                & 0.7241               & 0.6652                & 0.7184            & 0.5832           \\
mnasnet0\_5                                                                                                                                                                                                                                                                                          & 0.7111           & 0.7567          & 0.6451              & 0.3883            & 0.6056       & 0.3371                & 0.7454               & 0.7047                & 0.7325            & 0.6252           \\
vit\_base\_patch16\_224                                                                                                                                                                                                                                                                              & 0.7409           & 0.7783          & 0.4940              & 0.5323            & 0.6614       & 0.4823                & 0.7522               & 0.6134                & 0.6889            & 0.6382           \\
vit\_base\_patch32\_224                                                                                                                                                                                                                                                                              & 0.7455           & 0.7200          & 0.4860              & 0.5504            & 0.6861       & 0.5143                & 0.7404               & 0.6303                & 0.6801            & 0.6392           \\
vit\_large\_patch16\_224                                                                                                                                                                                                                                                                             & 0.7536           & 0.7457          & 0.4448              & 0.5092            & 0.6862       & 0.5054                & 0.7404               & 0.5986                & 0.6774            & 0.6290           \\
vit\_small\_patch32\_224 \(\star \)                                                                                                                                                                                                                                                                             & 0.8032           & 0.7711          & 0.5493              & 0.5198            & 0.6794       & 0.4563                & 0.7608               & 0.5995                & 0.7242            & 0.6515           \\
deit3\_small\_patch16\_224                                                                                                                                                                                                                                                                           & 0.7506           & 0.7255          & 0.5040              & 0.5367            & 0.6589       & 0.3785                & 0.7120               & 0.5672                & 0.6659            & 0.6110           \\
vit\_base\_patch8\_224                                                                                                                                                                                                                                                                               & 0.7445           & 0.7589          & 0.4917              & 0.4793            & 0.7042       & 0.5431                & 0.7648               & 0.5666                & 0.7056            & 0.6399           \\
vit\_tiny\_patch16\_224                                                                                                                                                                                                                                                                              & 0.7265           & 0.7632          & 0.4537              & 0.5148            & 0.6902       & 0.4575                & 0.7088               & 0.5518                & 0.6913            & 0.6175           \\
vit\_small\_patch16\_224                                                                                                                                                                                                                                                                             & 0.7749           & 0.7795          & 0.5317              & 0.5121            & 0.7154       & 0.5237                & 0.7360               & 0.5843                & 0.6943            & 0.6502           \\
vit\_base\_patch16\_384                                                                                                                                                                                                                                                                              & 0.7545           & 0.7474          & 0.4930              & 0.5158            & 0.7490       & 0.5285                & 0.7419               & 0.6177                & 0.6962            & 0.6493           \\
vit\_tiny\_patch16\_384                                                                                                                                                                                                                                                                              & 0.7272           & 0.7344          & 0.4344              & 0.4558            & 0.7090       & 0.4121                & 0.6912               & 0.5184                & 0.7044            & 0.5985           \\
vit\_small\_patch32\_384                                                                                                                                                                                                                                                                             & 0.7441           & 0.7714          & 0.4859              & 0.5124            & 0.7221       & 0.4376                & 0.7451               & 0.5888                & 0.7182            & 0.6362           \\
vit\_small\_patch16\_384                                                                                                                                                                                                                                                                             & 0.6993           & 0.7722          & 0.5083              & 0.5007            & 0.7224       & 0.5032                & 0.7481               & 0.6260                & 0.6719            & 0.6391           \\
vit\_base\_patch32\_384                                                                                                                                                                                                                                                                              & 0.7647           & 0.7674          & 0.4963              & 0.5299            & 0.7036       & 0.5191                & 0.7773               & 0.6445                & 0.6812            & \textbf{0.6538}  \\ \hline
Average                                                                                                                                                                                                                                                                                              & 0.7446           & \textbf{0.7471} & 0.4908              & 0.4742            & 0.6512       & 0.4208                & 0.7379               & 0.5468                & 0.6995            &                  \\ \hline
\end{tabular}
}
\label{BTlarge4c_def-para}
\end{table}

\subsection{Ablation studies}
\label{abstu}
We have carried-out four different versions as ablative studies such as using a 1) simple version, 2) normalization with PCA, 3) with SMOTE only, and 4) with the combination of normalization, PCA, and SMOTE for brain tumor classification. Each version have been discussed in the subsequent sections.

\subsubsection{Brain tumor classification using simple version}
In this experiment, we simply provide the top-2 or top-3 pre-trained DL based extracted features to 9 different ML classifiers.
The experimental results shown in Table \ref{BTsmall_top3_simple}, \ref{BTlarg2c_top3_simpleversion}, \ref{BTlarge4c_dataset_sim_vers} indicate the simple version means without considering normalization, PCA, and SMOTE methods. In both binary class datasets, the $SVM_{RBF}$ and $SVM_{linear}$ classifiers outperform all other ML classifiers as illustrated in Table \ref{BTsmall_top3_simple}, \ref{BTlarg2c_top3_simpleversion}, whereas, in multi-class dataset, the KNN classifier achieve the highest accuracy.

\begin{table}[!ht]
\centering
\caption{Accuracies of top three pre-trained models using simple version on BT-small-2c dataset. }
\scalebox{0.5}{
\begin{tabular}{lllllllllll}
\hline
\textbf{\begin{tabular}[c]{@{}l@{}}Deep Feature from the\\ Pre-Trained CNN and                                                                                                                                                                       \\ ViT Model\end{tabular}} & \multicolumn{10}{c}{\textbf{Machine Learning Classifier Accuracy}}                                                                                                                                                                                                                                                                                                                                             \\ \hline
\multicolumn{1}{c}{\textbf{}}                                                                                                                                                                                                                                                   & \multicolumn{1}{c}{\textbf{XGBoost}} & \multicolumn{1}{c}{\textbf{MLP}} & \multicolumn{1}{c}{\textbf{GaussianNB}} & \multicolumn{1}{c}{\textbf{Adaboost}} & \multicolumn{1}{c}{\textbf{KNN}} & \multicolumn{1}{c}{\textbf{RFClassifier}} & \multicolumn{1}{c}{\textbf{SVM\_linear}} & \multicolumn{1}{c}{\textbf{SVM\_sigmoid}} & \multicolumn{1}{c}{\textbf{SVM\_RBF}} & \multicolumn{1}{c}{\textbf{Average}} \\ \hline
vit\_small\_patch16\_224   + vit\_base\_patch16\_224                                                                                                                                                                                                                            & 0.9589                               & 0.9750                           & 0.8944                                  & 0.9250                                & 0.8694                           & 0.9750                                    & 0.9750                                   & 0.9750                                    & 0.9750                                & 0.9470                               \\ \hline
vit\_small\_patch32\_224 +   vit\_small\_patch16\_224                                                                                                                                                                                                                           & 0.9427                               & 0.9500                           & 0.9105                                  & 0.9589                                & 0.8782                           & 0.9589                                    & 0.9750                                   & 0.9750                                    & 0.9750                                & 0.9471                               \\ \hline
vit\_base\_patch16\_224 +   vit\_small\_patch32\_224                                                                                                                                                                                                                            & 0.9750                               & 0.9750                           & 0.9355                                  & 0.8677                                & 0.9266                           & 0.9500                                    & 0.9750                                   & 0.9589                                    & 0.9750                                & 0.9487                               \\ \hline
\begin{tabular}[c]{@{}l@{}}vit\_small\_patch16\_224 +   vit\_base\_patch16\_224 + \\ vit\_small\_patch32\_224\end{tabular}                                                                                                                                                      & 0.9750                               & 0.9750                           & 0.9194                                  & 0.9750                                & 0.8944                           & 0.9500                                    & 0.9750                                   & 0.9750                                    & 0.9750                                & \textbf{0.9571}                      \\ \hline
Average                                                                                                                                                                                                                                                                         & 0.9629                               & 0.9688                           & 0.9149                                  & 0.9317                                & 0.8921                           & 0.9585                                    & \textbf{0.9750}                          & 0.9710                                    & \textbf{0.9750}                       &                                      \\ \hline
\end{tabular}
}
\label{BTsmall_top3_simple}
\end{table}

\begin{table}[!ht]
\centering
\caption{Accuracies of top three pre-trained models using simple version on BT-large-2c dataset }
\scalebox{0.5}{
\begin{tabular}{ccccccccccc}
\hline
\multicolumn{1}{l}{\textbf{\begin{tabular}[c]{@{}l@{}}Deep Feature from the \\ Pre-Trained CNN and                                                                                                                                                                       \\ ViT Model\end{tabular}}} & \multicolumn{10}{c}{\textbf{Machine Learning Classifier Accuracy}}                                                                                                                                     \\ \cline{2-11} 
\textbf{}                                                                                                                                                                                                                                                                                            & \textbf{XGBoost} & \textbf{MLP} & \textbf{GaussianNB} & \textbf{Adaboost} & \textbf{KNN} & \textbf{RFClassifier} & \textbf{SVM\_linear} & \textbf{SVM\_sigmoid} & \textbf{SVM\_RBF} & \textbf{Average} \\ \hline
vit\_large\_patch16\_224   + vit\_base\_patch32\_384                                                                                                                                                                                                                                                 & 0.9933           & 0.995        & 0.88                & 0.99              & 0.99         & 0.9867                & 0.9967               & 0.9833                & 0.9983            & 0.9793           \\
vit\_large\_patch16\_224 +  vit\_small\_patch32\_384                                                                                                                                                                                                                                                 & 0.9883           & 0.9967       & 0.9                 & 0.9917            & 0.9883       & 0.99                  & 0.9967               & 0.985                 & 0.9983            & \textbf{0.9817}  \\
vit\_base\_patch32\_384 +   vit\_small\_patch32\_384                                                                                                                                                                                                                                                 & 0.99             & 0.9917       & 0.9133              & 0.99              & 0.9867       & 0.985                 & 0.9867               & 0.97                  & 0.995             & 0.9787           \\
\begin{tabular}[c]{@{}c@{}}vit\_large\_patch16\_224 + vit\_base\_patch32\_384\\ + vit\_small\_patch32\_384\end{tabular}                                                                                                                                                                              & 0.99             & 0.9967       & 0.8933              & 0.9917            & 0.9867       & 0.9917                & 0.9983               & 0.9833                & 0.9967            & 0.9809           \\ \hline
Average                                                                                                                                                                                                                                                                                              & 0.9904           & 0.995        & 0.8967              & 0.9908            & 0.9879       & 0.9883                & 0.9946               & 0.9804                & \textbf{0.9971}   &                  \\ \hline
\end{tabular}
}
\label{BTlarg2c_top3_simpleversion}
\end{table}

\begin{table}[!ht]
\centering
\caption{Accuracies of top 3 pre-trained models using simple version for BT-large 4c dataset}
\scalebox{0.5}{
\begin{tabular}{ccccccccccc}
\hline
\multicolumn{1}{l}{\textbf{\begin{tabular}[c]{@{}l@{}}Deep Feature from the \\ Pre-Trained CNN and                                                                                                                                                                       \\ ViT Model\end{tabular}}} & \multicolumn{10}{c}{\textbf{Machine Learning Classifier Accuracy}}                                                                                                                                        \\ \cline{2-11} 
\textbf{}                                                                                                                                                                                                                                                                                            & \textbf{XGBoost} & \textbf{MLP} & \textbf{GaussianNB} & \textbf{Adaboost} & \textbf{KNN}    & \textbf{RFClassifier} & \textbf{SVM\_linear} & \textbf{SVM\_sigmoid} & \textbf{SVM\_RBF} & \textbf{Average} \\ \hline
vit\_small\_patch32\_224   + mnasnet0\_5                                                                                                                                                                                                                                                             & 0.6947           & 0.7678       & 0.6593              & 0.7738            & 0.8286          & 0.7832                & 0.7847               & 0.7026                & 0.7703            & \textbf{0.7517}  \\
vit\_small\_patch32\_224 + vgg16                                                                                                                                                                                                                                                                     & 0.7066           & 0.7789       & 0.5856              & 0.7446            & 0.8072          & 0.7564                & 0.7767               & 0.6267                & 0.7814            & 0.7293           \\
mnasnet0\_5 + vgg16                                                                                                                                                                                                                                                                                  & 0.6403           & 0.7637       & 0.6447              & 0.7457            & 0.7819          & 0.7497                & 0.7586               & 0.6594                & 0.7591            & 0.7226           \\
\begin{tabular}[c]{@{}c@{}}vit\_small\_patch32\_224 + mnasnet0\_5 +\\    vgg16\end{tabular}                                                                                                                                                                                                          & 0.6989           & 0.7629       & 0.6331              & 0.7537            & 0.7954          & 0.7442                & 0.7691               & 0.6657                & 0.7548            & 0.7309           \\ \hline
Average                                                                                                                                                                                                                                                                                              & 0.6851           & 0.7683       & 0.6307              & 0.7545            & \textbf{0.8033} & 0.7584                & 0.7723               & 0.6636                & 0.7664            &                  \\ \hline
\end{tabular}
}
\label{BTlarge4c_dataset_sim_vers}
\end{table}

\subsubsection{Brain tumor classification using normalization and PCA}
Normalization is a feature scaling technique that standardizes features and minimizes variance, adjusting data to a scale between 0 and 1. Various feature scaling methods, such as Min-Max normalization \cite{rajagopal2020stacking}, Z-Score normalization \cite{henderi2021comparison}, and Decimal Scaling \cite{singh2022feature}, are employed across different studies. In this research, we applied the Min-Max normalization method to transform the data into a normalized format. This technique performs a linear transformation on the original dataset, preserving the relationships among the initial data values. It constrains the data to a range of [0, 1], and the Min-Max value is computed using equation \cite{han2022data}.

\begin{equation}
y_n=\left|\frac{y_0-y_{\min }}{y_{\max }-y_{\min }}\right|
\end{equation}

Here, $y_n$ represents the normalized data, $y_{min}$ refers to the smallest value in the dataset, $y_{max}$ signifies the largest value in the dataset, and $y_0$ is a specific value selected from the dataset.

In addition, Principal Component Analysis (PCA) is a technique for reducing the dimensionality of a dataset and selecting key features.  This technique ensures minimal loss of information from the input variables by using the fewest possible principal components. PCA converts a large set of features into a smaller group of principal components, aiming to preserve as much of the original dataset’s information as feasible. The benefits of PCA include eliminating correlated features that slow down processing, as well as handling missing variables and redundant data. It leverages multicollinearity to strip away such features from large datasets, thereby enhancing performance by cutting down training time through the removal of less impactful features. Additionally, PCA mitigates overfitting by trimming unnecessary features from the dataset. Here, we retain top 50\% of components for faster computation. 
Table \ref{BTsnorpca}, \ref{BT-large2c_top_three_with_norm_and_pca}, \ref{BT-lar-2c_with_nor_and_pca} illustrates the effect of normalization and PCA.

\begin{table}[!ht]
\centering
\caption{Accuracies of top three pre-trained models using normalization and PCA on BT-small-2c dataset. }
\scalebox{0.5}{
\begin{tabular}{ccccccccccc}
\hline
\multicolumn{1}{l}{\textbf{\begin{tabular}[c]{@{}l@{}}Deep Feature from the \\ Pre-Trained CNN and                                                                                                                                                                       \\ ViT Model\end{tabular}}} & \multicolumn{10}{c}{\textbf{Machine Learning Classifier Accuracy}}                                                                                                                                     \\ \cline{2-11} 
\textbf{}                                                                                                                                                                                                                                                                                            & \textbf{XGBoost} & \textbf{MLP} & \textbf{GaussianNB} & \textbf{Adaboost} & \textbf{KNN} & \textbf{RFClassifier} & \textbf{SVM\_linear} & \textbf{SVM\_sigmoid} & \textbf{SVM\_RBF} & \textbf{Average} \\ \hline
vit\_small\_patch16\_224   + vit\_base\_patch16\_224                                                                                                                                                                                                                                                 & 0.9339           & 0.9589       & 0.5806              & 0.8589            & 0.8694       & 0.7839                & 0.9750               & 0.9750                & 0.9750            & 0.8789           \\
vit\_small\_patch32\_224 +   vit\_small\_patch16\_224                                                                                                                                                                                                                                                & 0.9266           & 0.9839       & 0.7766              & 0.9000            & 0.8782       & 0.9089                & 0.9750               & 0.9589                & 0.9750            & \textbf{0.9203}  \\
vit\_base\_patch16\_224 +   vit\_small\_patch32\_224                                                                                                                                                                                                                                                 & 0.9750           & 0.9589       & 0.6129              & 0.9589            & 0.9177       & 0.9500                & 0.9750               & 0.9589                & 0.9750            & \textbf{0.9203}  \\
\begin{tabular}[c]{@{}c@{}}vit\_small\_patch16\_224 +   vit\_base\_patch16\_224 +\\  vit\_small\_patch32\_224\end{tabular}                                                                                                                                                                           & 0.9500           & 0.9589       & 0.5427              & 0.9589            & 0.9266       & 0.8855                & 0.9750               & 0.9750                & 0.9750            & 0.9053           \\ \hline
Average                                                                                                                                                                                                                                                                                              & 0.9464           & 0.9651       & 0.6282              & 0.9192            & 0.8980       & 0.8821                & 0.9750               & 0.9669                & \textbf{0.9750}   &                  \\ \hline
\end{tabular}
}
\label{BTsnorpca}
\end{table}

\begin{table}[!ht]
\centering
\caption{Accuracies of top three pre-trained models using normalization and pca on BT-large-2c dataset }
\scalebox{0.5}{
\begin{tabular}{lllllllllll}
\hline
\textbf{\begin{tabular}[c]{@{}l@{}}Deep Feature from the \\ Pre-Trained CNN and                                                                                                                                                                       \\ ViT Model\end{tabular}} & \multicolumn{10}{c}{\textbf{Machine Learning Classifier Accuracy}}                                                                                                                                                                                                                                                                                                                                             \\ \cline{2-11} 
\multicolumn{1}{c}{\textbf{}}                                                                                                                                                                                                                                                    & \multicolumn{1}{c}{\textbf{XGBoost}} & \multicolumn{1}{c}{\textbf{MLP}} & \multicolumn{1}{c}{\textbf{GaussianNB}} & \multicolumn{1}{c}{\textbf{Adaboost}} & \multicolumn{1}{c}{\textbf{KNN}} & \multicolumn{1}{c}{\textbf{RFClassifier}} & \multicolumn{1}{c}{\textbf{SVM\_linear}} & \multicolumn{1}{c}{\textbf{SVM\_sigmoid}} & \multicolumn{1}{c}{\textbf{SVM\_RBF}} & \multicolumn{1}{c}{\textbf{Average}} \\ \hline
vit\_large\_patch16\_224   + vit\_base\_patch32\_384                                                                                                                                                                                                                             & 0.9883                               & 0.9967                           & 0.665                                   & 0.9917                                & 0.9883                           & 0.9917                                    & 0.9967                                   & 0.975                                     & 0.9967                                & 0.9544                               \\
vit\_large\_patch16\_224 +   vit\_small\_patch32\_384                                                                                                                                                                                                                            & 0.9917                               & 0.9967                           & 0.6883                                  & 0.9933                                & 0.9883                           & 0.9867                                    & 0.995                                    & 0.97                                      & 0.9967                                & 0.9563                               \\
vit\_base\_patch32\_384 +   vit\_small\_patch32\_384                                                                                                                                                                                                                             & 0.985                                & 0.995                            & 0.7133                                  & 0.9917                                & 0.9917                           & 0.985                                     & 0.99                                     & 0.9783                                    & 0.995                                 & \textbf{0.9583}                      \\
\begin{tabular}[c]{@{}l@{}}vit\_large\_patch16\_224 +   vit\_base\_patch32\_384\\  + vit\_small\_patch32\_384\end{tabular}                                                                                                                                                       & 0.99                                 & 0.9983                           & 0.6767                                  & 0.995                                 & 0.985                            & 0.9933                                    & 0.9967                                   & 0.975                                     & 0.9967                                & 0.9563                               \\ \hline
Average                                                                                                                                                                                                                                                                          & 0.9888                               & \textbf{0.9967}                  & 0.6858                                  & 0.9929                                & 0.9883                           & 0.9892                                    & 0.9946                                   & 0.9746                                    & 0.9963                                &                                      \\ \hline
\end{tabular}
}
\label{BT-large2c_top_three_with_norm_and_pca}
\end{table}

\begin{table}[!ht]
\centering
\caption{Accuracies of top 3 pre-trained models using normalization and pca version for BT-large 4c dataset}
\scalebox{0.5}{
\begin{tabular}{ccccccccccc}
\hline
\multicolumn{1}{l}{\textbf{\begin{tabular}[c]{@{}l@{}}Deep Feature from the \\ Pre-Trained CNN and                                                                                                                                                                       \\ ViT Model\end{tabular}}} & \multicolumn{10}{c}{\textbf{Machine Learning Classifier Accuracy}}                                                                                                                                        \\ \cline{2-11} 
\textbf{}                                                                                                                                                                                                                                                                                            & \textbf{XGBoost} & \textbf{MLP} & \textbf{GaussianNB} & \textbf{Adaboost} & \textbf{KNN}    & \textbf{RFClassifier} & \textbf{SVM\_linear} & \textbf{SVM\_sigmoid} & \textbf{SVM\_RBF} & \textbf{Average} \\ \hline
vit\_small\_patch32\_224   + mnasnet0\_5                                                                                                                                                                                                                                                             & 0.5206           & 0.7455       & 0.4412              & 0.7553            & 0.8244          & 0.7717                & 0.7419               & 0.6975                & 0.7581            & \textbf{0.6951}  \\
vit\_small\_patch32\_224 + vgg16                                                                                                                                                                                                                                                                     & 0.5024           & 0.7542       & 0.3420              & 0.7263            & 0.7342          & 0.7830                & 0.7788               & 0.7208                & 0.7559            & 0.6775           \\
mnasnet0\_5 + vgg16                                                                                                                                                                                                                                                                                  & 0.4601           & 0.7658       & 0.4157              & 0.7294            & 0.7967          & 0.7781                & 0.7382               & 0.6933                & 0.7540            & 0.6813           \\
vit\_small\_patch32\_224 +   mnasnet0\_5 + vgg16                                                                                                                                                                                                                                                     & 0.5083           & 0.7619       & 0.4178              & 0.7193            & 0.8042          & 0.7618                & 0.7407               & 0.7080                & 0.7540            & 0.6862           \\ \hline
Average                                                                                                                                                                                                                                                                                              & 0.4978           & 0.7569       & 0.4042              & 0.7326            & \textbf{0.7898} & 0.7737                & 0.7499               & 0.7049                & 0.7555            &                  \\ \hline
\end{tabular}
}
\label{BT-lar-2c_with_nor_and_pca}
\end{table}

\subsubsection{Brain tumor classification using SMOTE only}
 ML techniques face difficulties when a single class heavily dominates a dataset, indicating that the instances of that class greatly outnumber those of other classes. This type of dataset is known as imbalanced, and such disparity can distort classification processes and undermine the reliability of the outcomes. To counter this problem, SMOTE, an oversampling method introduced by Chawla et al. \cite{chawla2002smote}, is utilized. SMOTE synthetically increases the minority class by creating artificial samples using the KNN technique \cite{chawla2002smote}, thus equilibrating the dataset. The SMOTE algorithm is implemented in diverse fields to address imbalance issues, such as network intrusion detection systems \cite{cieslak2006combating}, breast cancer diagnosis \cite{fallahi2011expert}, and detecting sentence boundaries in speech \cite{saha2010awareness}. While SMOTE is extensively employed in numerous fields, it has certain limitations, including overgeneralization and high variance \cite{vida2016practical}. SMOTE method synthetically maximizes the minority class using the following equation:

 \begin{equation}
    x_{s y n}=x_i+\left(x_{k n n}-x_i\right) \times t
 \end{equation}
 
The following steps described the SMOTE.

\begin{itemize}
    \item  Determine the feature vector $x_i$ and find its k-nearest neighbors $x_{knn}$.
    \item Computes the difference between the K-NN and feature vector.
\item Scale the difference by a random number between 0 and 1.
\item Incorporates the output number into the feature vector to determine a new point on the line segment.
\item Repeats steps 1 through 4 to determine the feature vectors.
\end{itemize}

Similarly, in this work, we utilized the synthetic minority oversampling approach to address the issues of overfitting and imbalanced samples by balancing these significantly imbalanced MRI datasets. Table \ref{BT-small-2c-top3-using_smot-only}, \ref{BTlarge2c_smote_only}, \ref{Bt-larg-4c_smot_only} shows the results of oversampling methods i.e., SMOTE.

\begin{table}[!ht]
\centering
\caption{Accuracies of top three pre-trained models using smote data on BT-small-2c dataset}
\scalebox{0.5}{
\begin{tabular}{ccccccccccc}
\hline
\multicolumn{1}{l}{\textbf{\begin{tabular}[c]{@{}l@{}}Deep Feature from the \\ Pre-Trained CNN and                                                                                                                                                                       \\ ViT Model\end{tabular}}} & \multicolumn{10}{c}{\textbf{Machine Learning Classifier Accuracy}}                                                                                                                                     \\ \cline{2-11} 
\textbf{}                                                                                                                                                                                                                                                                                            & \textbf{XGBoost} & \textbf{MLP} & \textbf{GaussianNB} & \textbf{Adaboost} & \textbf{KNN} & \textbf{RFClassifier} & \textbf{SVM\_linear} & \textbf{SVM\_sigmoid} & \textbf{SVM\_RBF} & \textbf{Average} \\ \hline
vit\_small\_patch16\_224   + vit\_base\_patch16\_224                                                                                                                                                                                                                                                 & 0.9750           & 0.9750       & 0.9266              & 0.9500            & 0.8548       & 0.9758                & 0.9755               & 0.9355                & 0.9757            & 0.9460           \\
vit\_small\_patch32\_224 +   vit\_small\_patch16\_224                                                                                                                                                                                                                                                & 0.9589           & 0.9750       & 0.8444              & 0.9589            & 0.8871       & 0.9339                & 0.9757               & 0.9105                & 0.9758            & 0.9306           \\
vit\_base\_patch16\_224 +   vit\_small\_patch32\_224                                                                                                                                                                                                                                                 & 0.9750           & 0.9750       & 0.9194              & 0.9750            & 0.8548       & 0.9751                & 0.9753               & 0.9266                & 0.9750            & \textbf{0.9470}  \\
\begin{tabular}[c]{@{}c@{}}vit\_small\_patch16\_224 +   vit\_base\_patch16\_224 +\\  vit\_small\_patch32\_224\end{tabular}                                                                                                                                                                           & 0.9750           & 0.9750       & 0.9355              & 0.9750            & 0.8710       & 0.9750                & 0.9754               & 0.8944                & 0.9750            & \textbf{0.9470}  \\ \hline
Average                                                                                                                                                                                                                                                                                              & 0.9710           & 0.9750       & 0.9065              & 0.9647            & 0.8669       & 0.9650                & \textbf{0.9755}      & 0.9168                & 0.9754            &                  \\ \hline
\end{tabular}
}
\label{BT-small-2c-top3-using_smot-only}
\end{table}

\begin{table}[!ht]
\centering
\caption{Accuracies of top three pre-trained models using smote only for BT-large 2c dataset}
\scalebox{0.5}{
\begin{tabular}{ccccccccccc}
\hline
\multicolumn{1}{l}{\textbf{\begin{tabular}[c]{@{}l@{}}Deep Feature from the \\ Pre-Trained CNN and                                                                                                                                                                       \\ ViT Model\end{tabular}}} & \multicolumn{10}{c}{\textbf{Machine Learning Classifier Accuracy}}                                                                                                                                     \\ \cline{2-11} 
\textbf{}                                                                                                                                                                                                                                                                                            & \textbf{XGBoost} & \textbf{MLP} & \textbf{GaussianNB} & \textbf{Adaboost} & \textbf{KNN} & \textbf{RFClassifier} & \textbf{SVM\_linear} & \textbf{SVM\_sigmoid} & \textbf{SVM\_RBF} & \textbf{Average} \\ \hline
vit\_large\_patch16\_224   + vit\_base\_patch32\_384                                                                                                                                                                                                                                                 & 0.9917           & 0.9967       & 0.88                & 0.9917            & 0.99         & 0.9867                & 0.9967               & 0.985                 & 0.9983            & 0.9796           \\
vit\_large\_patch16\_224 +   vit\_small\_patch32\_384                                                                                                                                                                                                                                                & 0.99             & 0.9983       & 0.9017              & 0.99              & 0.9883       & 0.9917                & 0.9967               & 0.985                 & 0.9983            & \textbf{0.9822}  \\
vit\_base\_patch32\_384 +   vit\_small\_patch32\_384                                                                                                                                                                                                                                                 & 0.9867           & 0.995        & 0.9133              & 0.9917            & 0.9867       & 0.9883                & 0.9867               & 0.97                  & 0.995             & 0.9793           \\
\begin{tabular}[c]{@{}c@{}}vit\_large\_patch16\_224 +   vit\_base\_patch32\_384\\  + vit\_small\_patch32\_384\end{tabular}                                                                                                                                                                           & 0.9917           & 0.995        & 0.8933              & 0.995             & 0.9867       & 0.9883                & 0.9983               & 0.9833                & 0.9967            & 0.9809           \\ \hline
Average                                                                                                                                                                                                                                                                                              & 0.99             & 0.9963       & 0.8971              & 0.9921            & 0.9879       & 0.9887                & 0.9946               & 0.9808                & \textbf{0.9971}   &                  \\ \hline
\end{tabular}
}
\label{BTlarge2c_smote_only}
\end{table}

\begin{table}[!ht]
\centering
\caption{Accuracies of top 3 pre-trained models using smote only version for BT-large 4c dataset}
\scalebox{0.5}{
\begin{tabular}{ccccccccccc}
\hline
\multicolumn{1}{l}{\textbf{\begin{tabular}[c]{@{}l@{}}Deep Feature from the \\ Pre-Trained CNN and                                                                                                                                                                       \\ ViT Model\end{tabular}}} & \multicolumn{10}{c}{\textbf{Machine Learning Classifier Accuracy}}                                                                                                                                        \\ \cline{2-11} 
\textbf{}                                                                                                                                                                                                                                                                                            & \textbf{XGBoost} & \textbf{MLP} & \textbf{GaussianNB} & \textbf{Adaboost} & \textbf{KNN}    & \textbf{RFClassifier} & \textbf{SVM\_linear} & \textbf{SVM\_sigmoid} & \textbf{SVM\_RBF} & \textbf{Average} \\ \hline
vit\_small\_patch32\_224   + mnasnet0\_5                                                                                                                                                                                                                                                             & 0.6566           & 0.7639       & 0.6593              & 0.7738            & 0.8286          & 0.7832                & 0.7847               & 0.7026                & 0.7703            & \textbf{0.7470}  \\
vit\_small\_patch32\_224 + vgg16                                                                                                                                                                                                                                                                     & 0.6395           & 0.7892       & 0.5856              & 0.7446            & 0.8072          & 0.7564                & 0.7767               & 0.6267                & 0.7814            & 0.7230           \\
mnasnet0\_5 + vgg16                                                                                                                                                                                                                                                                                  & 0.5975           & 0.7579       & 0.6447              & 0.7457            & 0.7819          & 0.7497                & 0.7586               & 0.6594                & 0.7591            & 0.7172           \\
vit\_small\_patch32\_224 + mnasnet0\_5 +   vgg16                                                                                                                                                                                                                                                     & 0.6443           & 0.7569       & 0.6331              & 0.7537            & 0.7954          & 0.7442                & 0.7691               & 0.6657                & 0.7548            & 0.7241           \\ \hline
Average                                                                                                                                                                                                                                                                                              & 0.6345           & 0.7670       & 0.6307              & 0.7545            & \textbf{0.8033} & 0.7584                & 0.7723               & 0.6636                & 0.7664            &                  \\ \hline
\end{tabular}
}
\label{Bt-larg-4c_smot_only}
\end{table}

\subsubsection{Brain tumor classification using normalization and PCA along with SMOTE}
In this section, we integrate all three techniques to evaluate their combined benefits. Table \ref{BTlarge_2c_norm_and_pca_smot}, \ref{BTswith_norm_and_PCA_smotE}, \ref{bt-large-4c-with-norma-and-pca-with-smote} shows the results of normalizing the data, reducing the dimensionality using PCA, and oversampling using SMOTE. As shown in Table \ref{BTlarge_2c_norm_and_pca_smot} and \ref{BTswith_norm_and_PCA_smotE} the $SVM_{RBF}$ outperforms all other ML classifiers in the 2-class scenario. In contrast, Table \ref{bt-large-4c-with-norma-and-pca-with-smote} demonstrates that KNN achieves superior performance in the 4-class scenario.

\begin{table}[!ht]
\centering
\caption{Accuracies of top 3 pre-trained models using normalization and pca along with smote for BT large 2c dataset}
\scalebox{0.5}{
\begin{tabular}{ccccccccccc}
\hline
\multicolumn{1}{l}{\textbf{\begin{tabular}[c]{@{}l@{}}Deep Feature from the \\ Pre-Trained CNN and                                                                                                                                                                       \\ ViT Model\end{tabular}}} & \multicolumn{10}{c}{\textbf{Machine Learning Classifier Accuracy}}                                                                                                                                     \\ \cline{2-11} 
\textbf{}                                                                                                                                                                                                                                                                                            & \textbf{XGBoost} & \textbf{MLP} & \textbf{GaussianNB} & \textbf{Adaboost} & \textbf{KNN} & \textbf{RFClassifier} & \textbf{SVM\_linear} & \textbf{SVM\_sigmoid} & \textbf{SVM\_RBF} & \textbf{Average} \\ \hline
vit\_large\_patch16\_224   + vit\_base\_patch32\_384                                                                                                                                                                                                                                                 & 0.9883           & 0.9967       & 0.6667              & 0.9917            & 0.9883       & 0.9933                & 0.9967               & 0.97                  & 0.9967            & 0.9543           \\
vit\_large\_patch16\_224 +   vit\_small\_patch32\_384                                                                                                                                                                                                                                                & 0.99             & 0.9983       & 0.6983              & 0.9933            & 0.9883       & 0.985                 & 0.995                & 0.965                 & 0.9967            & 0.9567           \\
vit\_base\_patch32\_384 +   vit\_small\_patch32\_384                                                                                                                                                                                                                                                 & 0.985            & 0.9933       & 0.725               & 0.99              & 0.9917       & 0.9883                & 0.99                 & 0.9783                & 0.995             & \textbf{0.9596}  \\
\begin{tabular}[c]{@{}c@{}}vit\_large\_patch16\_224 +   vit\_base\_patch32\_384\\  + vit\_small\_patch32\_384\end{tabular}                                                                                                                                                                           & 0.9933           & 0.995        & 0.6783              & 0.9933            & 0.9867       & 0.9917                & 0.9983               & 0.98                  & 0.9967            & 0.957            \\ \hline
Average                                                                                                                                                                                                                                                                                              & 0.9892           & 0.9958       & 0.6921              & 0.9921            & 0.9887       & 0.9896                & 0.995                & 0.9733                & \textbf{0.9963}   &                  \\ \hline
\end{tabular}
}
\label{BTlarge_2c_norm_and_pca_smot}
\end{table}

\begin{table}[]
\centering
\caption{Accuracies of top three pre-trained models using normalization and PCA along with SMOTE on BT-small-2c dataset }
\scalebox{0.5}{
\begin{tabular}{lllllllllll}
\hline
\textbf{\begin{tabular}[c]{@{}l@{}}Deep Feature from the \\ Pre-Trained CNN and                                                                                                                                                                       \\ ViT Model\end{tabular}} & \multicolumn{10}{c}{\textbf{Machine Learning Classifier Accuracy}}                                                                                                                                                                                                                                                                                                                                             \\ \cline{2-11} 
\multicolumn{1}{c}{\textbf{}}                                                                                                                                                                                                                                                    & \multicolumn{1}{c}{\textbf{XGBoost}} & \multicolumn{1}{c}{\textbf{MLP}} & \multicolumn{1}{c}{\textbf{GaussianNB}} & \multicolumn{1}{c}{\textbf{Adaboost}} & \multicolumn{1}{c}{\textbf{KNN}} & \multicolumn{1}{c}{\textbf{RFClassifier}} & \multicolumn{1}{c}{\textbf{SVM\_linear}} & \multicolumn{1}{c}{\textbf{SVM\_sigmoid}} & \multicolumn{1}{c}{\textbf{SVM\_RBF}} & \multicolumn{1}{c}{\textbf{Average}} \\ \hline
vit\_small\_patch16\_224   + vit\_base\_patch16\_224                                                                                                                                                                                                                             & 0.9339                               & 0.975                            & 0.6177                                  & 0.8839                                & 0.9032                           & 0.8589                                    & 0.975                                    & 0.975                                     & 0.975                                 & 0.8997                               \\
vit\_small\_patch32\_224 +   vit\_small\_patch16\_224                                                                                                                                                                                                                            & 0.9105                               & 1                                & 0.6016                                  & 0.9177                                & 0.871                            & 0.8839                                    & 0.975                                    & 0.9427                                    & 0.975                                 & 0.8975                               \\
vit\_base\_patch16\_224 +   vit\_small\_patch32\_224                                                                                                                                                                                                                             & 0.9589                               & 0.9589                           & 0.6355                                  & 0.9839                                & 0.846                            & 0.9                                       & 0.975                                    & 0.9427                                    & 0.975                                 & \textbf{0.9084}                      \\
\begin{tabular}[c]{@{}l@{}}vit\_small\_patch16\_224 +   vit\_base\_patch16\_224 + \\ vit\_small\_patch32\_224\end{tabular}                                                                                                                                                       & 0.9589                               & 0.9589                           & 0.5177                                  & 0.8855                                & 0.8871                           & 0.925                                     & 0.975                                    & 0.9427                                    & 0.975                                 & 0.8918                               \\ \hline
Average                                                                                                                                                                                                                                                                          & 0.9405                               & 0.9732                           & 0.5931                                  & 0.9177                                & 0.8768                           & 0.8919                                    & \textbf{0.975}                           & 0.9508                                    & \textbf{0.975}                        &                                      \\ \hline
\end{tabular}
}
\label{BTswith_norm_and_PCA_smotE}
\end{table}

\begin{table}[!ht]
\centering
\caption{Accuracies of top 3 pre-trained models using normalization and pca with smote version for BT-large 4c dataset}
\scalebox{0.5}{
\begin{tabular}{ccccccccccc}
\hline
\multicolumn{1}{l}{\textbf{\begin{tabular}[c]{@{}l@{}}Deep Feature from the \\ Pre-Trained CNN and                                                                                                                                                                       \\ ViT Model\end{tabular}}} & \multicolumn{10}{c}{\textbf{Machine Learning Classifier Accuracy}}                                                                                                                                        \\ \cline{2-11} 
\textbf{}                                                                                                                                                                                                                                                                                            & \textbf{XGBoost} & \textbf{MLP} & \textbf{GaussianNB} & \textbf{Adaboost} & \textbf{KNN}    & \textbf{RFClassifier} & \textbf{SVM\_linear} & \textbf{SVM\_sigmoid} & \textbf{SVM\_RBF} & \textbf{Average} \\ \hline
vit\_small\_patch32\_224   + mnasnet0\_5                                                                                                                                                                                                                                                             & 0.5340           & 0.7453       & 0.4369              & 0.7502            & 0.8391          & 0.7846                & 0.7408               & 0.7037                & 0.7451            & \textbf{0.6977}  \\
vit\_small\_patch32\_224 + vgg16                                                                                                                                                                                                                                                                     & 0.5039           & 0.7479       & 0.3420              & 0.7280            & 0.7292          & 0.7600                & 0.7738               & 0.7164                & 0.7509            & 0.6725           \\
mnasnet0\_5 + vgg16                                                                                                                                                                                                                                                                                  & 0.4865           & 0.7260       & 0.4135              & 0.7034            & 0.7992          & 0.7591                & 0.7609               & 0.6893                & 0.7440            & 0.6758           \\
vit\_small\_patch32\_224 + mnasnet0\_5 +   vgg16                                                                                                                                                                                                                                                     & 0.5017           & 0.7522       & 0.4157              & 0.7253            & 0.8117          & 0.7725                & 0.7550               & 0.6947                & 0.7590            & 0.6875           \\ \hline
Average                                                                                                                                                                                                                                                                                              & 0.5065           & 0.7429       & 0.4020              & 0.7267            & \textbf{0.7948} & 0.7691                & 0.7576               & 0.7010                & 0.7497            &                  \\ \hline
\end{tabular}
}
\label{bt-large-4c-with-norma-and-pca-with-smote}
\end{table}

\subsection{Ensemble of Classifiers}
\label{ecl}
The ensemble method aims to improve the generalization and dependability of a classifier by integrating various conceptually unique base ML classifiers into a single, unified classifier \cite{goodfellow2016deep}. The goal is to develop a combined classifier that outperforms any of the individual base classifiers on its own. The outcomes of the fourth set of experiments, where we considered single improved version (i.e., the combination of normalization, PCA, and SMOTE), which employed the top-2 or top-3 ML classifier, are presented in Table \ref{ML-ensemble-2c-small}, \ref{ensemble_ML-BT-large2c}, and \ref{4c_ML_ensemble}. The primary objective of ensembling ML classifiers is to further improve the brain tumor classification accuracy. The tables indicate that the performance of the ensembled ML models significantly improved compared to the outcomes of the prior three experiments. Moreover, this ensemble technique is advantageous as it combines the strengths of various classifiers, resulting in more reliable predictions.

\begin{table}[!ht]
\centering
\caption{Accuracies of ensembled ML classifiers with normalization, PCA, and SMOTE on BT-small-2c dataset.}
\scalebox{0.5}{
\begin{tabular}{ccccccc}
\hline
\multicolumn{1}{l}{\textbf{\begin{tabular}[c]{@{}l@{}}Machine Learning \\ensembling \\                                                                                                                                                                        \end{tabular}}} & \multicolumn{6}{c}{\textbf{Pre-trained DL ensembling models }}                                                                                                                                                                                                                                               \\ \cline{2-7} 
\textbf{}                                                                                                                                                                                                                                                                                            & \multicolumn{1}{l}{\textbf{vit\_base\_patch16\_224}} & \multicolumn{1}{l}{\textbf{vit\_small\_patch16\_224}} & \multicolumn{1}{l}{\textbf{vit\_small\_patch32\_224}} & \multicolumn{1}{l}{\textbf{vit\_base\_patch16\_384}} & \multicolumn{1}{l}{\textbf{vit\_tiny\_patch16\_224}} & \multicolumn{1}{l}{\textbf{Average}} \\ \hline
SVM\_linear +   SVM\_RBF                                                                                                                                                                                                                                                                             & 0.9750                                               & 0.9500                                                & 0.9750                                                & 0.9750                                               & 0.9250                                               & 0.9600                               \\
SVM\_linear + MLP                                                                                                                                                                                                                                                                                    & 0.9750                                               & 0.9500                                                & 0.9750                                                & 0.9750                                               & 0.9250                                               & 0.9600                               \\
SVM\_RBF + MLP                                                                                                                                                                                                                                                                                       & 0.9500                                               & 0.9750                                                & 0.9750                                                & 0.9750                                               & 0.9750                                               & \textbf{0.9700}                      \\
SVM\_linear + SVM\_RBF + MLP                                                                                                                                                                                                                                                                         & 0.9750                                               & 0.9750                                                & 0.9750                                                & 0.9589                                               & 0.9500                                               & 0.9668                               \\ \hline
Average                                                                                                                                                                                                                                                                                              & 0.9688                                               & 0.9625                                                & \textbf{0.9750}                                       & 0.9710                                               & 0.9438                                               &                                      \\ \hline
\end{tabular}
}
\label{ML-ensemble-2c-small}
\end{table}

\begin{table}[!ht]
\centering
\caption{accuracies of ensembled ML classifiers with normalization, PCA, and SMOTE on BT-large-2c dataset}
\scalebox{0.5}{
\begin{tabular}{ccccccc}
\hline
\multicolumn{1}{l}{\textbf{\begin{tabular}[c]{@{}l@{}}Machine Learning \\ensembling \\                                                                                                                                                                       \end{tabular}}} & \multicolumn{6}{c}{\textbf{Pre-trained DL ensembling models}}                                                                                                                       \\ \cline{2-7} 
\textbf{}                                                                                                                                                                                                                                                                                            & \textbf{vit\_large\_patch16\_224} & \textbf{vit\_base\_patch32\_384} & \textbf{vit\_small\_patch32\_384} & \textbf{vit\_base\_patch32\_224} & \textbf{vit\_base\_patch16\_384} & \textbf{Average} \\ \hline
KNN + MLP                                                                                                                                                                                                                                                                                            & 0.9900                            & 0.9917                           & 0.9867                            & 0.9917                           & 0.9867                           & 0.9894           \\
KNN + SVM\_RBF                                                                                                                                                                                                                                                                                       & 0.9900                            & 0.9933                           & 0.9950                            & 0.9933                           & 0.9850                           & 0.9913           \\
MLP + SVM\_RBF                                                                                                                                                                                                                                                                                       & 0.9950                            & 0.9900                           & 0.9933                            & 0.9967                           & 0.9867                           & 0.9923           \\
KNN + SVM\_RBF + MLP                                                                                                                                                                                                                                                                                 & 0.9950                            & 0.9933                           & 0.9950                            & 0.9967                           & 0.9917                           & \textbf{0.9943}  \\ \hline
Average                                                                                                                                                                                                                                                                                              & 0.9925                            & 0.9921                           & 0.9925                            & \textbf{0.9946}                  & 0.9875                           &                  \\ \hline
\end{tabular}
}
\label{ensemble_ML-BT-large2c}
\end{table}

\begin{table}[!ht]
\centering
\caption{Accuracies of ensembled ML classifiers with normalization, PCA, and SMOTE on BT-large-4c dataset.}
\scalebox{0.6}{
\begin{tabular}{ccccccc}
\hline
\multicolumn{1}{l}{\textbf{\begin{tabular}[c]{@{}l@{}}Machine Learning \\ensembling \\                                                                                                                                                                       \end{tabular}}} & \multicolumn{6}{c}{\textbf{Pre-trained DL ensembling models}}                                                                                         \\ \cline{2-7} 
\textbf{}                                                                                                                                                                                                                                                                                            & \textbf{vit\_small\_patch32\_224} & \textbf{mnasnet0\_5} & \textbf{vgg16} & \textbf{vit\_base\_patch32\_384} & \textbf{vit\_small\_patch16\_224} & \textbf{Average} \\ \hline
RFClassifier + MLP                                                                                                                                                                                                                                                                                   & 0.98                              & 0.9417               & 0.96           & 0.9883                           & 0.9867                            & 0.97134          \\
RFClassifier + KNN                                                                                                                                                                                                                                                                                   & 0.9867                            & 0.9317               & 0.96           & 0.99                             & 0.9817                            & 0.97002          \\
MLP + KNN                                                                                                                                                                                                                                                                                            & 0.9917                            & 0.9633               & 0.9783         & 0.9933                           & 0.9867                            & \textbf{0.98266} \\
RFClassifier + MLP + KNN                                                                                                                                                                                                                                                                             & 0.9917                            & 0.96                 & 0.9717         & 0.9933                           & 0.9917                            & 0.98168          \\ \hline
Average                                                                                                                                                                                                                                                                                              & 0.987525                          & 0.949175             & 0.9675         & \textbf{0.991225}                & 0.9867                            &                  \\ \hline
\end{tabular}
}
\label{4c_ML_ensemble}
\end{table}

\section{Discussion} \label{discussion}
In the discussion section of our study, we emphasize the importance of our proposed approach for classifying brain tumors. By integrating image preprocessing and augmentation, sophisticated feature extraction, feature selection, and ML models, we achieved exceptional outcomes in classifying brain tumor. The use of pre-processing as shown in section \ref{pp}, significantly improved image quality and reduced noise, resulting in more dependable findings. Additionally, feature importance analysis and HPO enhanced the model’s effectiveness, achieving an outstanding accuracy for each classifier. To compare our fine tune hyperparameter ML models as shown in Table \ref{BTs2call}, \ref{BTlarge2c}, and \ref{BTlarge4c_hypertune} with default hyperparameter of ML classifier in Table \ref{BT-small-2c-default-param}, \ref{BTlarg2c_defaultpara}, and \ref{BTlarge4c_def-para} demonstrating the advantage of our method in accurately identifying brain tumor.

Moreover, the ensembling strategies both feature ensembling and ML classifier ensembling contribute to further improvements in accuracy, thereby substantiating the effectiveness of our proposed methodology.

To evaluate the significance of image pre-processing, we conducted an experiment on the BT-large-2c dataset without performing any pre-processing step, using only pre-trained CNN models as shown in Table \ref{nopreprocessing} and achieved poor results as compare to pre-processed BT-large-2c dataset as illustrated in Table \ref{BTlarg2c_defaultpara}. 

Our work underscores the potential of ML to assist clinicians in delivering accurate and trustworthy brain tumor diagnoses, paving the way for practical clinical applications and better patient outcomes.


\section{Conclusion and Future Work} \label{con}
This study presents a comprehensive and effective approach for classifying brain tumors, incorporating image preprocessing and augmentation, feature extraction, feature selection, and ML models. The proposed approach, which fuses deep features, significantly simplifies distinguishing between brain cancer cases and and normal cases. By identifying the most critical features and optimizing the model's performance through hyperparameter tuning, we found that MLP was the best-performing ML model with $vit\_base\_patch16\_224$ deep feature and is a good choice with high accuracy on a small-size binary class dataset. Whereas, the ensemble of $vit\_large\_patch16\_224 +   vit\_base\_patch32\_384\\  + vit\_small\_patch32\_384$ deep models with $SVM_{RBF}$ classifier is an accurate choice in two class classification with a large size dataset. On the other hand, the ensemble of ML classifiers such as $RF + MLP + KNN$ is a good choice where there are four classes.

In summary, our proposed feature ensemble effectively addresses the limitations of individual DL models, while the proposed machine learning classifier ensemble mitigates the drawbacks of single classifiers. Together, these approaches deliver better and more robust performance, particularly for big datasets. These findings suggest that our dual ensembling strategy is well-suited for brain tumor classification. However, while the performance of our method is encouraging, future research is necessary to optimize model size for deployment in real-time medical diagnosis systems, potentially through the use of knowledge distillation techniques. Additionally, due to its strong generalization capability, this approach can be expanded to address other classification challenges in the medical domain and serves as a practical tool for classification tasks in various other fields.

\section*{CRediT authorship contribution statement}
\textbf{Zahid Ullah:} Conceptualization, Data curation, Methodology, Software, Formal analysis, Investigation, Writing - original draft, Writing - review \& editing.  \textbf{Dragan Pamucar:} Conceptualization, Writing – review \& editing, Formal analysis, Supervision.  \textbf{Jihie Kim:} Formal analysis, Investigation, Supervision, Project administration.

\section*{\textbf{Declaration of Competing Interests}} The authors declare that they have no known competing financial interests or personal relationships that could have appeared to influence the work reported in this paper.

\section*{Acknowledgements}
This research was supported by the MSIT(Ministry of Science and ICT), Korea, under the ITRC(Information Technology Research Center) support program(IITP-2025-RS-2020-II201789), and the Artificial Intelligence Convergence Innovation Human Resources Development(IITP-2025-RS-2023-00254592) supervised by the IITP(Institute for Information \& Communications Technology Planning \& Evaluation).

\bibliographystyle{elsarticle-num-names}
\bibliography{sample.bib}







\end{document}